\documentclass{article}

\usepackage{microtype}
\usepackage{graphicx}
\usepackage{subcaption}
\usepackage{booktabs} % for professional tables

\usepackage{hyperref}

\usepackage[preprint]{icml2026}

\usepackage{amsmath}
\usepackage{amssymb}
\usepackage{mathtools}
\usepackage{amsthm}

\usepackage[utf8]{inputenc} % allow utf-8 input
\usepackage[T1]{fontenc}    % use 8-bit T1 fonts
\usepackage{hyperref}       % hyperlinks
\usepackage{url}            % simple URL typesetting
\usepackage{booktabs}       % professional-quality tables
\usepackage{amsfonts}       % blackboard math symbols
\usepackage{nicefrac}       % compact symbols for 1/2, etc.
\usepackage{microtype}      % microtypography
\usepackage{xcolor}% colors

\usepackage{float}
\usepackage{amsmath}
\usepackage{caption}
\usepackage{subcaption}
\usepackage{graphicx}
\usepackage{url}

\def\RR{\mathbb{R}}
\def\EE{\mathbb{E}}
\def\PP{\mathbb{P}}

\def\OO{\mathcal{O}}
\def\BB{\mathcal{B}}
\def\FF{\mathcal{F}}
\def\GG{\mathcal{G}}

\usepackage[capitalize,noabbrev]{cleveref}

\theoremstyle{plain}
\newtheorem{theorem}{Theorem}[section]

\newtheorem{lemma}[theorem]{Lemma}
\newtheorem{corollary}[theorem]{Corollary}
\theoremstyle{definition}

\newtheorem{assumption}[theorem]{Assumption}
\theoremstyle{remark}

\usepackage[textsize=tiny]{todonotes}

\icmltitlerunning{Divergence Results and Convergence of a Variance Reduced Version of Adam}

\begin{document}

\twocolumn[
  \icmltitle{Divergence Results and Convergence of a Variance Reduced Version of Adam}

  % It is OKAY to include author information, even for blind submissions: the
  % style file will automatically remove it for you unless you've provided
  % the [accepted] option to the icml2026 package.

  % List of affiliations: The first argument should be a (short) identifier you
  % will use later to specify author affiliations Academic affiliations
  % should list Department, University, City, Region, Country Industry
  % affiliations should list Company, City, Region, Country

  % You can specify symbols, otherwise they are numbered in order. Ideally, you
  % should not use this facility. Affiliations will be numbered in order of
  % appearance and this is the preferred way.
  \icmlsetsymbol{equal}{*}

  \begin{icmlauthorlist}
    \icmlauthor{Ruiqi Wang}{1}
    \icmlauthor{Diego Klabjan}{1}
    
  \end{icmlauthorlist}

  \icmlaffiliation{1}{ Industrial Engineering and Management Sciences, Northwestern University, Evanston, IL, USA, 60201}
  
  \icmlcorrespondingauthor{Ruiqi Wang}{RuiqiWang2025@u.northwestern.edu}
  % \icmlcorrespondingauthor{Diego Klabjan}{d-klabjan@northwestern.edu}

  % You may provide any keywords that you find helpful for describing your
  % paper; these are used to populate the "keywords" metadata in the PDF but
  % will not be shown in the document
  \icmlkeywords{Machine Learning, ICML}

  \vskip 0.3in
]

% this must go after the closing bracket ] following \twocolumn[ ...

% This command actually creates the footnote in the first column listing the
% affiliations and the copyright notice. The command takes one argument, which
% is text to display at the start of the footnote. The \icmlEqualContribution
% command is standard text for equal contribution. Remove it (just {}) if you
% do not need this facility.

% Use ONE of the following lines. DO NOT remove the command.
% If you have no special notice, KEEP empty braces:
\printAffiliationsAndNotice{}  % no special notice (required even if empty)
% Or, if applicable, use the standard equal contribution text:
% \printAffiliationsAndNotice{\icmlEqualContribution}

\begin{abstract}
  Stochastic optimization algorithms using exponential moving averages of the past gradients, such as Adam, RMSProp and AdaGrad, have been having great successes in many applications, especially in training deep neural networks. Adam in particular stands out as efficient and robust. Despite of its outstanding performance, Adam has been proved to be divergent for some specific problems. We revisit the divergent question and provide divergent examples under stronger conditions such as in expectation or high probability. Under a variance reduction assumption, we show that an Adam-type algorithm converges, which means that it is the variance of gradients that causes the divergence of original Adam. To this end, we propose a variance reduced version of Adam and provide a convergent analysis of the algorithm. Numerical experiments demonstrate that the proposed algorithm outperforms Adam in both training and validation performances. From the perspective of experimental efficiency, we introduce an online version of our algorithm that eliminates the need for full gradient computations. By trading off a small degree of accuracy in the full gradient approximation, we demonstrate by means of experiments that the algorithm is also suitable for large language models.
\end{abstract}

\section{Introduction}
Stochastic optimization based on mini-batch is a common training procedure in machine learning. Suppose we have finitely many differentiable objectives $\{f_{n}(w)\}_{n=1}^{N}$ defined on $\RR^d$ with $N$ being the size of the training set. In each iteration, a random index set $\BB_t$ is selected from $\{1,\ldots,N\}$ and the update is made based on the mini-batch loss $F^{\BB_t}(w) = \frac{1}{b}\sum_{n\in\BB_t} f_n(w)$, where $b=|\BB_t|$ is the batch size. The goal is to minimize the empirical risk $\min_{w\in\RR^d} F(w) := \frac{1}{N}\sum_{n=1}^N f_{n}(w).$

First order methods, which make updates based on the information of the gradient of mini-batch loss functions, prevail in practice \citep{Goodfellow-et-al-2016}. A simple method is stochastic gradient descent (SGD), where the model parameters are updated at the negative direction of the mini-batch loss gradient in each iteration. Several adaptive variants of SGD, such as AdaGrad \citep{duchi2011adaptive}, RMSProp \citep{hinton2012neural} and Adam \citep{kingma2014adam}, are proved to converge faster than SGD in practice. These methods take the historical gradients into account. A general Adam framework is introduced in Algorithm~\ref{alg:Adam}, where vanilla Adam takes a mini-batch gradient as the gradient estimation $G(w_t;\xi_t)$. 

\begin{algorithm}[tb]
\caption{General Adam}\label{alg:Adam}
\begin{algorithmic}
\REQUIRE Gradient estimation $\GG(\cdot;\cdot)$, seed generation rule $\PP_\xi$, initial point $w_1$, hyper parameters $\alpha_t$, $0\leq \beta_1, \beta_2<1$, 
 and $\epsilon>0$.
\STATE $m_0\gets 0$, $v_0\gets 0$
\FOR{$t\in 1,\ldots T$}
\STATE Sample $g_t \gets \GG(w_t;\xi_t)$ where $\xi_t\sim\PP_\xi$
\STATE $m_t \gets \beta_1 m_{t-1} + (1-\beta_1) g_t$
\STATE $v_t \gets \beta_2 v_{t-1} + (1-\beta_2) g_t \odot g_t$
\STATE $V_t \gets \mathrm{diag}(v_t) + \epsilon I_d$
\STATE $w_{t+1} \gets w_t - \alpha_t V_t^{-1/2} m_t$
\ENDFOR
\end{algorithmic}
\end{algorithm}
Although Adam achieves great success in many tasks, it may fail to solve some problems. \citep{reddi2019convergence} found a flaw in the proof of convergence in \citep{kingma2014adam} and proposed a divergent example for online Adam. Based on the divergent example, they pointed out that when some large, informative but rare gradients occur, the exponential moving average would make them decay quickly and hence would lead to the failure of convergence. To this end, Reddi et al. proposed two variants of Adam to fix this problem. The first proposal, known as AMSGrad, suggests taking the historical maximum of the Adam state $v_t$ in order to obtain `long-term memories' and prevent the large and informative gradients from being forgotten. Although this helps keeping the information of large gradients, it hurts the adaptability of Adam. If the algorithm is exposed to a large gradient at early iterations, the $v_t$ parameter will stay constant, hence the algorithm will not automatically adapt the step size, and it will degenerate to a momentum method. Another intuitive criticism is that keeping $v_t$ increasing is against what one expects, since if the algorithm converges, the norm of gradients should decrease and $v_{t+1} - v_t = (1-\beta_2) (g_t^2-v_t)$ is more likely to be negative, where $g_t$ is the stochastic gradient in step $t$ and $\beta_2$ is a hyper parameter.

Several other proposals tried to fix the divergent problem of Adam. The second variant proposed in \citep{reddi2019convergence}, called AdamNC, requires the second order moment hyper-parameter $\beta_2$ to increase and to satisfy several conditions. However the conditions are hard to check. Although they claim that $\beta_{2,t}=1-1/t$ satisfies the conditions, this case is actually AdaGrad, which is already well-known for its convergence. \citep{zhou2018adashift} analyzed the divergent example in \citep{reddi2019convergence}, and pointed out that the correlation of $v_t$ and $g_t$ causes divergence of Adam, and proposed a decorrelated variant of Adam. The theoretical analysis in \citep{zhou2018adashift} is based on complex assumptions and they do not provide a convergence analysis of their algorithm. Several other works, such as \citep{guo2021novel, shi2020rmsprop, wang2019sadam, zou2019sufficient} suggested properly tuning the hyper-parameters of Adam-type algorithms had helped with convergence in practice.

It is empirically well-known that larger batch size reduces the variance of the loss of a stochastic optimization algorithm. \citep{qian2020impact} gave a theoretical proof that the variance of the stochastic gradient is proportional to $1/b$. Although several works connected the convergence of Adam with the mini-batch size, the direct connection between convergence and variance is wanted. For the full-batch case (i.e., where there is no variance), \citep{de2018convergence} showed that Adam converges under some specific scheduling of learning rates. \citep{shi2020rmsprop} showed the convergence of full gradient Adam and RMSProp with the learning rate schedule $\alpha_t = \alpha / \sqrt{t}$ and constants $\beta_1$ and $\beta_2$ satisifying $\beta_1<\sqrt{\beta_2}$. For the stochastic setting with a fixed batch size, \citep{zaheer2018adaptive} proved that the expected norm of the gradient can be bounded into a neighborhood of 0, whose size is proportional to $1/b$. They suggested to increase the batch size with the number of iterations in order to establish convergence. One question is that whether there exists a threshold of batch size $b^*<N$, such that any batch size larger than $b^*$ guarantees convergence. We show that even when $b=N-1$, there still exist divergent examples of Adam. This means that although large batch size helps tighten the optimality gap, the convergence issue is not solved as long as the variance exists. Another possible convergent result is to analyze the convergence in expectation or high probability under a stochastic starting point. However our divergent result holds for any initial point, which rules out this possibility.

Without relying on the mini-batch size, we make a direct analysis of variance and the convergence of Adam. We first show a motivating result which points out that the convergence of an Adam-type algorithm can be implied by reducing the variance. Motivated by this, we propose a variance reduced version of Adam, called VRADAM, and show that VRADAM converges. We provide two options regarding to resetting of Adam states during the full gradient steps, and recommend the resetting option based on a theoretical analysis herein and computational experiments. Computing the full gradient can be computationally expensive for large-scale datasets. To address this, we propose an efficient version of online VRADAM that replaces the full gradient with the historical mean of snapshot gradients. This mean is dynamically updated and converges to the true full gradient over time. Theoretically, we demonstrate that this approach introduces an additional error term, which is bounded by $\OO(1/m)$, where $m$ represents the number of iterations between two updates of the snapshot model. Through extensive experiments, we show that VRADAM outperforms existing algorithms, such as ADAM, for various tasks and datasets. Additionally, we demonstrate that the online version of VRADAM achieves superior performance compared to ADAM in fine-tuning large language models (LLMs).

In Section 3, we show a divergent example. Using contradiction by assuming the algorithm converges, we show that the expected update of iterates is larger than a positive constant, which means that it is impossible for the algorithm to converge to an optimal solution, which contradicts with the assumption. In Section 5, we prove the convergence of VRADAM. The main proof technique applied is to properly bound the difference between the estimated gradients and the true value of gradients. By bounding the update of the objective function in each iterate, we can further employ the strong convexity assumption and conclude convergence.

Our contributions are as follows: 1. We provide an unconstrained and strongly convex stochastic optimization problem on which the original Adam diverges. We show that the divergence holds for any initial point, which rules out all of the possible weaker convergent results under stochastic starting point. 2. We construct a divergent mini-batch problem with $b=N-1$, and conclude that there does not exist a convergent threshold for the mini-batch size. 3. We propose a variance reduced version of Adam. We provide convergence results of the variance reduced version for strongly convex objectives to optimality or non-convex objectives. We show by experiments that the variance reduction does not harm the numerical performance of Adam. 4. We propose a computationally efficient version of VRADAM that avoids full gradient computations and demonstrates superior performance over ADAM in the task of fine-tuning LLMs. We propose a version of Adam that converges and exhibits superior practical performances as the original Adam.

In Section 2, we review the literature on the topics of the convergence/divergence issue of Adam and variance reduction optimization methods. In Section 3 we provide divergent examples for stochastic Adam. We show that the example is divergent for large batch sizes, which disproves the existence of a convergence threshold of mini-batch size. In Section 4 we start from a reducing variance condition and prove the convergence of an Adam-type algorithm under this condition. In Section 5 we propose a variance reduced version of Adam. We also provide a convergence result of our variance reduced Adam. In Section 6 we conduct several numerical experiments and show the convergence and sensitivity of the proposed algorithm.
\section{Literature Review}
\textbf{Convergence of Adam:} \citep{reddi2019convergence} firstly pointed out the convergence issue of Adam and proposed two convergent variants: (a) AMSGrad takes the historical maximum value of $v_t$ to keep the step size decreasing and (b) AdamNC requires the hyper-parameters to satisfy specific conditions. Both of the approaches require that $\beta_1$ varies with time, which is inconsistent with practice. \citep{fang2019convergence} gave a convergence proof for AMSGrad with constant $\beta_1$ and \citep{alacaoglu2020new} provided a tighter bound. Enlarging the mini-batch size is another direction. \citep{de2018convergence} and \citep{shi2020rmsprop} proved the convergence of Adam for full batch gradients and \citep{zaheer2018adaptive} showed the convergence of Adam as long as the batch size is of the same order as the maximum number of iterations, but one criticism is that such a setting for the batch size is very inefficient in practice since the calculation of a large batch gradient is expensive. Several works, such as \citep{guo2021novel, zou2019sufficient, wang2019sadam} proposed guidelines on setting hyper-parameters in order to obtain convergent results.  \citep{guo2021novel} showed that as long as $\beta_1$ is close enough to 1, in particular, $1-\beta_{1, t}\propto{1/\sqrt{t}}$, Adam establishes a convergent rate of $\OO(1/\sqrt{T})$. However, since \citep{reddi2019convergence} proposed the divergent example for any fixed $\beta_1$ and $\beta_2$ such that $\beta_1<\sqrt{\beta_2}$, the proof cannot hold for constant momentum parameters. \citep{zou2019sufficient} also provided a series of conditions under which Adam could converge. Specifically, they require the quantity $\alpha_t/\sqrt{1-\beta_{2, t}}$ to be `almost' non-increasing. \citep{wang2019sadam} proposed to set the denominator hyper-parameter $\epsilon$ to be $1/t$, and showed the convergence of Adam for strongly convex objectives. The aforementioned works focus on setting the hyper-parameters in Adam. On contrary, our work proposes a new algorithm that only requires basic and common conditions. We show a $\OO(T^{-p})$ convergence rate for $0<p<1$ where $p$ is dependent on hyper-parameters. 

Recent work focuses on the impact of $\beta_1$ and $\beta_2$ on the convergence performance.
\citep{zhang2022adam} proposed the convergence region for the values of $(\beta_1,\beta_2)$, showing that Adam converges to the neighborhood of critical points. \citep{li2023convergence} provided analysis of Adam when $\beta_1$ converges to 0. \citep{iiduka2022theoretical} studied on the case without the Lipschitz continuous condition and concluded that Adam performs better when $\beta_1$ and $\beta_2$ are close to 1. According to \citep{iiduka2022theoretical} and practical selections for $\beta_1$ and $\beta_2$, our work focuses on the case when they are constants. \citep{zhang2023adamconvergemodificationupdate} show that the norm of gradients for vanilla Adam is bounded by $\OO (\log T/\sqrt{T}+\sqrt{D_0})$ where the second term vanishes when $\beta_2\to 1$. However, the coefficient of the first term goes to infinity as $\beta_2\to 1$.

\textbf{Variance reduction:} The computational efficiency issues of full gradient descent methods get more severe with a large data size, but employing stochastic gradient descent may cause divergence because of the issue of variance. One classic method for variance reduction is to use mini-batch losses with a larger batch size, which however does not guarantee the variance to converge to zero. As an estimation of the full gradient, the stochastic average gradient (SAG) method \citep{le2013stochastic} uses an average of $\nabla f_i(x_{k_i})$, where $k_i$ is the most recent step index when sample $i$ is picked. Although the convergence analysis of SAG provided in \citep{schmidt2017minimizing} showed its remarkable linear convergence, the estimator of the descent direction is biased and the analysis of SAG is complicated. SAGA \citep{defazio2014saga}, an unbiased variant to SAG introduced a concept called `covariates' and guarantees linear convergence as well. Both SAG and SAGA require the memory of $\OO(Nd)$, which is expensive when the data set is large. SVRG \citep{johnson2013accelerating} constructs two layers of iterations and calculates the full gradient as an auxiliary vector for variance reduction before starting each inner loop. It only requires a memory of $\OO(d)$. Most of the literature on variance reduction focus on the convergence rate and memory requirement on the plain SGD algorithm. Recently, \citep{dubois2021svrg} combined AdaGrad with SVRG for robustness in the learning rate. \citep{huang2021superadam} proposed an adaptive learning rate framework, SUPER-Adam, which is able to integrate variance reduction techniques. SUPER-Adam applies an inner second order optimization with the adaptive matrix, which enhances the convergence rate of $\sqrt{\log(T)}/T^{1/3}$. \citep{liu2020Adam} propose a modified version of Adam and show that the algorithm enjoys adaptive variance reduction. However the convergence result is crucially dependent on the assumption that $\sum_t^T \|m_t\|_2\leq T^\alpha$  for some $\alpha < 1$, which is not required in our work. Our work introduces the idea of variance reduction to the convergence analysis of Adam. It initiates the idea of the dynamic learning rate to SVRG.
\section{Divergent Examples for Stochastic Adam with Large Batch Size}
\label{sec:divergent}
Several recent works \citep{shi2020rmsprop, zaheer2018adaptive, de2018convergence} have suggested increasing the mini-batch size may help with convergence of Adam. In particular, vanilla Adam is convergent if the mini-batch size $b$ is equal to the size of the training set, or it increases in the same order as training iterates. An interesting question is whether there exists a threshold of the batch size $b^*=b(N)$, which is smaller than $N$, such that $b>b^*$ implies convergence of Adam. If such a threshold exists, the convergence can be guaranteed by a sufficiently large, but neither increasing nor as large as the training set size, batch size. Unfortunately, such a threshold does not exist. In fact, we show in this section that as long as the algorithm is not full batch, one can find a divergent example of Adam.

Another aspect of interest is if Adam converges on average or with high probability. Our example establishes non-convergence for any initial data point (even starting with an optimal one). We conclude that a probabilistic statement is impossible if stochasticity comes from either sampling or the initial point.

Reddi et al. firstly proposed a divergent example for Adam in \citep{reddi2019convergence}. The example, which is under the population loss minimization framework, consists of two linear functions defined on a finite interval. One drawback of this example is that the optimization problem is constrained, yet training in machine learning is usually an unconstrained problem. Under the unconstrained framework, the example proposed in \citep{reddi2019convergence} does not have a minimum solution, hence it does not satisfy the basic requirements. We firstly propose an unconstrained problem under the population loss minimization framework.

Let a random variable $\xi$ take discrete value from the set $\{1,2\}$, and set $\PP(\xi = 1) = \frac{1+\delta}{1+\delta^4}$ for some $\delta>1$. Furthermore, we define the estimation of gradients by $ \GG(w;1)= \frac{w}{\delta} + \delta^4$ and $\GG(w;2)= \frac{w}{\delta} - 1$, which implies the stochastic optimization problem with the loss functions $f_1(w) = \frac{w^2}{2\delta} + \delta^4 w$ and $f_2(w) = \frac{w^2}{2\delta} - w$ with the corresponded probability distribution with respect to $\xi$. The population loss is given as $F(w) = \EE_{\xi}\left[ f_\xi(w)\right]$.
We call this stochastic optimization problem the \textbf{Original Problem}($\delta$), or \textbf{OP}($\delta$) for short. The optimal solution of OP($\delta$) is $w^* = -\delta^2$. We should note that OP($\delta$) is defined on $\RR$, thus it is unconstrained. In addition, it is a strongly convex problem. As a divergent property of OP($\delta$), we show the following result.

\begin{theorem}
\label{thm:DivOP}
There exists a $\delta^*>2$ such that for any $\delta>\delta^*$ and any initial point $w_1$, with learning rate $\alpha_t$ satisfying $\sum_{t=1}^\infty \alpha_t =+\infty$ and $\alpha_t\leq \alpha$, Adam does not converge in expectation on OP($\delta$), i.e., $\EE[F(w_t)]\not \to F^*$ where $F^*$ is the optimal value of $F(w)$.
\end{theorem}
We remark that for OP($\delta$), the decreasing learning rate of Adam also leads to non-convergence. The proof of Theorem\ref{thm:DivOP} is given in the appendix, where we show that for large enough $\delta$, assuming the convergence $\EE[F(w_t)]\to F^*$ implies that the expectation of the Adam update between two consequential iterates is always positive. As a consequence, the iterates keep drifting from the optimal solution, and hence leads to contradiction. The divergent example also tells us that strong convexity and the relaxation of constraints cannot help with the convergence of Adam.

Based on the construction of OP($\delta$), we can give the divergent examples for any fixed mini-batch size.
\begin{theorem}
\label{thm:DivFixb}
For any fixed $b$, there exists an $N^*_b$, such that for any $N>N^*_b$, there exists a mini-batch problem with sample size $N$ and batch size $b$ where Adam does not converge for any initial point.
\end{theorem}
Even if the batch size is unreasonably large, say $b=N-1$, we can still construct the divergent example based on OP($\delta$) as stated next.
\begin{theorem}
\label{thm:DivLargeb}
 There exists an $N^*$ such that for any $N>N^*$, there exists a mini-batch problem with sample size $N$ and batch size $b=N-1$ where Adam does not converge for any initial point.
\end{theorem}
In conclusion, Theorem~\ref{thm:DivFixb} and Theorem~\ref{thm:DivLargeb} extinguish the hope of finding a large enough batch size for stochastic Adam to converge. Among the related works regarding the convergence of Adam and batch size, larger batch size is always suggested, but the results in this section have enlightened the limitations of such approaches. We display the numerical results of this example in the appendix.
\section{Motivation}

\label{sec:motivate}
In this section, we stick with the general Adam algorithm described in Algorithm~\ref{alg:Adam}. To analyze, we make several assumptions on the gradient estimator and objective.
\begin{assumption}
\label{assmpt:Basic}
The gradient estimator $\GG:\RR^d\times \Omega\to\RR^d$ and objective $F:\RR^d \to \RR$ satisfy the following: \textbf{(1)} $\GG$ is unbiased, i.e., for any $w\in \RR^d$, $\EE_\xi\left[\GG(w;\xi)\right] = \nabla F(w)$. \textbf{(2)} There exists a constant $0<L<+\infty$, such that for any $\xi\in \Omega$ and $w,\bar{w}\in \RR^d$, we have $\| \GG( w;\xi) - \GG(\bar w;\xi)\|_2\leq L \|w-\bar w\|_2$ and $\|\nabla F(w)- \nabla F(\bar w)\|_2\leq L\|w-\bar w\|_2$. \textbf{(3)} There exists a constant $0<G<+\infty$, such that for any $\xi\in\Omega$ and $w\in\RR^d$, we have $\|\GG(w;\xi)\|_2\leq G$ and $\|\nabla F(w)\|_2\leq G$.
\end{assumption}
At this point convexity is not needed. We mainly focus on the variance of the gradient estimator. The common assumptions in the literature are that the variance is bounded by a constant \citep{zaheer2018adaptive}, or a linear function of the square of the norm of the objective  $\mathrm{Var}(\GG_i(w;\xi)) \leq C_1 + C_2 \|\nabla F(w)\|_2^2$ \citep{bottou2018optimization}. Another assumption made in \citep{shi2020rmsprop, vaswani2019fast} is called the `strongly growth condition' which is $\sum_{n=1}^N \|\nabla f_n(w)\|_2^2\leq C \|\nabla F(w)\|_2^2$ for some $C>0$. Note that for vanilla Adam where $\GG(w;\xi)=\nabla F^{\BB}(w)$ the strongly growth condition implies that $\nabla F^{\BB}(w^*) = 0$ if and only if $\nabla F(w^*)=0$. As a result the strongly growth condition implies that $\mathrm{Var}(\GG(w;\xi))\leq 2L \EE[\|w-w^*\|_2^2]$,
given Lipshitz smooth gradients for full-batch and mini-batch losses. For those iterates close to a saddle point, the variance is automatically reduced, because $\|w-w^*\|_2^2$ is small. However, the strongly growth condition is so strong that the majority of practical problems do not satisfy it. In fact, one observation of OP($\delta$) is that the variance is a constant, which also breaks the strongly growth condition. 

In this section, as a motivative result, let us assume the variance of the gradient estimator is reduced a priori.
Let us denote a series of positive constants $\{\lambda_t\}_{t= 1}^T$ such that for any $t=1,\ldots, T$, we have $\mathrm{Var}(\GG (w_t; \xi_t))\leq \lambda_t.\nonumber$
For the objective with a finite lower bound, we have the following result.
\begin{theorem}
\label{thm:MtvNonCvx}
Let Assumption~\ref{assmpt:Basic} be satisfied, and assume that $F(w)$ is lower bounded by $F_{\mathrm{inf}}>-\infty$. Then for any $w_1$, Adam satisfies
$    \min_{1\leq t \leq T} \EE\left[\left\|\nabla F(w_t)\right\|_2^2\right] \leq \OO\left( \frac{\sum_{t=1}^T \alpha_t^2 + \alpha_t\lambda_t}{\sum_{t=1}^T \alpha_t}\right).\nonumber
$
\end{theorem}
The proof is in the appendix. Let us assume that the two common conditions $\sum_{t=1}^\infty \alpha_t = \infty$ and $\sum_{t=1}^\infty \alpha_t^2 <\infty$ are satisfied. Theorem~\ref{thm:MtvNonCvx} shows that Adam converges if $\sum_{t=1}^{\infty} \alpha_t\lambda_t<+\infty$. In fact, $\lambda_t \to 0 $ as $t\to\infty$ implies that $\sum_{t=1}^{T}\alpha_t \lambda_t / \sum_{t=1}^T \alpha_t \to 0$, and hence leads to convergence.

We emphasize that since the assumption on variance is made on the algorithmic iterates $\{w_t\}_{t=1}^T$, it is very difficult to be checked for a specific problem in advance. However, we showed that if the variance is convergent, an Adam-type algorithm converges. We show that the algorithm we propose has convergent variance and furthermore is convergent.
\section{Variance Reduced Adam}
\label{sec:VRADAM}

\begin{algorithm}[tb]
\caption{Variance Reduced Adam}\label{alg:VRADAM}
\begin{algorithmic}
\REQUIRE Loss functions $\{f_n(w)\}_{n=1}^N$, initial point $\widetilde w_1$, Adam hyper-parameters $\alpha_t$ , $0\leq \beta_1, \beta_2<1$ and $\epsilon>0$, mini-batch size $b$, projection parameters $M$ and $U<1$, inner iteration size $m$. Initialize $m_m^{(0)}\gets 0$, $v_m^{(0)}\gets 0$. $\widetilde w_1 \gets \Pi_{\mathbf{B}_M}\left(\widetilde w_1\right)$
\FOR{$t= 1,\ldots, T$}
\STATE $w_1^{(t)}\gets \tilde  w_t$
\STATE \textbf{(A)}: $m_0^{(t)}\gets 0$, $v_0^{(t)}\gets 0$,
\STATE \textbf{(B)}: $m_0^{(t)}\gets m_{m}^{(t-1)}$ and $v_0^{(t)}\gets v_{m}^{t-1}$
\FOR{$k=1,\ldots,m$}
\STATE Sample $\BB_k^{(t)}$ from $\{1,\ldots,N\}$ with $\left|\BB_k^{(t)}\right|=b$.
\STATE $g_k^{(t)}\gets \nabla F^{\BB_k^{(t)}}\left(w_k^{(t)}\right) - \nabla F^{\BB_k^{(t)}}(\widetilde w_t) + \nabla F(\widetilde w_t)$ 
\STATE $m_k^{(t)} \gets \beta_1 m_{k-1}^{(t)} + (1-\beta_1)g_k^{(t)}$, \;\;$v_k^{(t)} \gets \beta_2 v_{k-1}^{(t)} + (1-\beta_2)g_k^{(t)}\odot g_k^{(t)}$
\STATE  \textbf{(A)}: $\widetilde m^{(t)}_k \gets \frac{m^{(t)}_k}{1-\beta_1^k}$, $\widetilde v^{(t)}_k \gets \frac{v^{(t)}_k}{1-\beta_2^{k}}$, 
\STATE \textbf{(B)}: $\widetilde m^{(t)}_k \gets \frac{m^{(t)}_k} {1-\beta_1^{k+(t-1)m}}$, $\widetilde v^{(t)}_k \gets \frac{v^{(t)}_k}{1-\beta_2^{k+(t-1)m}}$
\STATE $V_k^{(t)}\gets \mathrm{diag}\left(\widetilde v_k^{(t)} + \epsilon\right)$, 
\STATE $w^{(t)}_{k+1} \gets w^{(t)}_k - \alpha_t \left(V_k^{(t)}\right)^{-1/2}\widetilde m_k^{(t)}$
\ENDFOR
\IF{$\left\|w_{m+1}^{(t)}\right\|_2 > M$}
\STATE $M' \gets \min\left\{M, U \left\|w_{m+1}^{(t)}\right\|_2\right\}$, \STATE $\widetilde w_{t+1}\gets \Pi_{\mathbf{B}_{M'}}\left(w_{m+1}^{(t)}\right)$
\ELSE 
\STATE{$\widetilde w_{t+1} \gets w_{m+1}^{(t)}$}
\ENDIF
\ENDFOR
\end{algorithmic}
\end{algorithm}

Variance reduction for random variables is a common topic in many fields. In general, an unbiased variance reduction of a random variable $X$ is $\tilde X = X - Y + \EE Y$, which establishes the variance $\mathrm{Var}(\tilde X) = \mathrm{Var}(X) + \mathrm{Var}(Y) - 2\mathrm{Cov}(X,Y)<\mathrm{Var}(X)\nonumber$ given $\mathrm{Cov}(X,Y)>\mathrm{Var}(Y)/2$, i.e., $X$ and $Y$ are positively correlated at a sufficient level. In the context of stochastic gradient descent, the random variable for variance reduction is $\GG(w_t;\xi_t)$, the gradient of mini-batch loss $\nabla F^{\BB_t}(w_t)$. \citep{johnson2013accelerating} proposed a solution for SGD. They suggested the associate random variable to be the gradient of the same mini-batch loss at a previous iterate $\tilde w$. Since the expectation of a mini-batch gradient is the full-batch gradient, the descent direction becomes $g_t = \nabla F^{\BB_t}(w_t) - \nabla F^{\BB_t}(\tilde w) + \nabla F(\tilde w)$. Vector $\tilde w$ is known as the snapshot model. Since calculation of the full batch gradient at $\tilde w$ is required, \citep{johnson2013accelerating} proposed to save the snapshot model every $m$ iterations, which is known as the SVRG algorithm. Inspired by SVRG and motivated by the result in Section~\ref{sec:motivate}, we propose the combination of the variance reduce method and Adam, called VRADAM (Algorithm~\ref{alg:VRADAM}). 
In Algorithm~\ref{alg:VRADAM}, we let $\mathbf{B}_{M'} = \{w\in\RR^d : \left\|w\right\|_2 \leq M'\}$ be the ball with radius $M'$ and let $\Pi_{\mathbf{B}_{M'}}(w) = \arg\min_{w'\in \mathbf{B}_{M'}} \left\|w-w'\right\|_2 $ be the projection operator to the ball.

Computing full gradients can be computationally expensive for certain large-scale datasets. To address this, we propose an online version of our approach that eliminates the need for full gradient computation in the outer loop. Instead, we approximate the full gradient using the historical average of snapshot models. The adjustment with Algorithm~\ref{alg:VRADAM} is in the updates of $g_k^{(t)}$ where we have $\frac{1}{k} \sum_{l=1}^k \nabla F^{\BB_l^{(t)}}(\widetilde w_t)$ replacing the full gradient $\nabla F(\widetilde w_t)$. We also provide two options: (A) and (B) with regard to the update of Adam states. In the appendix, we provide a theoretical analysis that recommends option A, the resetting option.

In the previous section we show the advantage of resetting of the Adam states every outer iteration over not doing so by comparing the values of the objectives in the two options. In this section, we provide a convergence proof of the resetting option of VRADAM. We show the convergence results of Adam for strongly convex and non-convex loss functions. 
 
\begin{assumption}
\label{assmpt:LSmoothGrad}
    \textbf{(1)} There exists a constant $0<L<\infty$, such that for any $n=1,\ldots,N$ and $w,\bar w\in\RR^d$, we have $\|\nabla f_n(w) -\nabla f_n(\bar w)\|_2 \leq L\|w-\bar w\|_2$. \textbf{(2)} There exists a constant $0<G<\infty$, such that for all $w\in\RR^d$ and $1\leq n\leq N$, we have $\left\|\nabla f_n(w)\right\|_2\leq G$. Furthermore, $F$ is lower bounded, i.e., $F_{\inf} := \inf_{w\in\RR^d} F(w) >-\infty$. \textbf{(3)}  $F(w)$ is $c$-strongly convex. Furthermore, letting $w^*$ be the unique global minimum of $F$, we assume that $M$ satisfies $\frac{\sqrt{L^2+c^2}}{c}\left\|w^*\right\|_2<M<+\infty$ and we set $U=\frac{c^2}{L^2+c^2}$.
\end{assumption}

We assume that each sample loss has a Lipschitz continuous gradient. For the non-convex case, we assume that the gradients of loss functions are upper bounded, and the empirical risk is lower bounded. For the strongly-convex case, the bounded gradient assumption contradicts strong convexity if the feasible region is $\RR^d$, since $\left\|\nabla F(w)\right\|_2^2=\left\|\nabla F(w)-\nabla F(w*)\right\|_2^2\geq c\left\|w-w^*\right\|_2^2$, where $w^*$ is the global minimum. We take a projection of the iterates to a bounded region $\mathbf{B}_{M'}$ at the end of each outer loop, as exhibited in Algorithm~\ref{alg:VRADAM}. We remark that in order to check the requirement of $M$ in Item 3 in Assumption~\ref{assmpt:LSmoothGrad}, one needs to know the value of $w^*$, which is unknown before solving the problem. However, we show in the appendix that for several common types of loss functions, such as mean square error, cross entropy and softened hinge, $\|w^*\|_2$ can be upper bounded by a function of input data with linear complexity, which means that our assumptions are practically testable. 

We start analyzing the strongly convex case. Although we do not require bounded gradients, the projection operation helps bounding the gradients of each sample loss at each iterate.
\begin{lemma}
    \label{lem:BoundGradient}
    Given (1) and (3) in Assumption~\ref{assmpt:LSmoothGrad}, and $\alpha_t\leq \alpha$ for all $t$, there exists $0<G<+\infty$ such that 
    $\left\|\nabla f_n\left(w_k^{(t)}\right)\right\|_2 \leq G\nonumber$ for all $n,t$ and $k$.
\end{lemma}

The proof is in the appendix. Given the result in Lemma~\ref{lem:BoundGradient}, we show the convergence result of VRADAM for strongly convex loss functions.
\begin{theorem}
\label{thm:VRStrCvx}
Let (1) and (3) in Assumption~\ref{assmpt:LSmoothGrad} be satisfied. Let $\alpha_t = \alpha/t$ and we require $C_2 = 2c(1-\beta_1)/\sqrt{9G^2 + \epsilon}< 1/\alpha m$, where $G$ is given in Lemma~\ref{lem:BoundGradient}. Then for any initial point $w_1$, VRADAM with Option (A) satisfies $ F(\widetilde w_T) - F^* \leq \OO\left( T^{-C_2m\alpha}\right)$ almost surely.
\end{theorem}
\color{black}
We remark that the requirement $C_2 m \alpha < 1$ can be satisfied by properly selected $\alpha$ and $\beta_1$. Specifically, when $\beta_1$ is close to $1$ and $\alpha$ is small, the assumption is more likely to be satisfied. The proof starts by bounding the update of the objective function in one iterate and then applies strong convexity. Regarding the objectives that are not necessarily convex, we define a random index $\tau$ sampled from $\{1,\ldots,T\}$, where $\PP(\tau = t) = \alpha_t /\sum_{t=1}^T \alpha_t$. We exhibit the following result.
\begin{theorem}
\label{thm:VRNonCvx}
Let (1) and (2) in Assumption~\ref{assmpt:LSmoothGrad} be satisfied and let $M=+\infty$. Then for any initial point $w_1$, VRADAM with Option (A) satisfies $\EE [\left\| \nabla F(\widetilde w_\tau)\right\|_2^2] \leq \OO\left(\sum_{t=1}^T \alpha_t^2 / \sum_{t=1}^T\alpha_t\right)$.
\end{theorem}
We remark that given $\sum_{t=1}^\infty \alpha_t =+\infty$ and $\sum_{t=1}^\infty \alpha_t^2 <+\infty$, $\EE [\left\| \nabla F(\widetilde w_\tau)\right\|_2^2]\to0$ as $T\to+\infty$. Under the general objective setting, as a corollary of Theorem~\ref{thm:VRNonCvx}, we can bound the variance of the norm of the gradient. Corollary~\ref{cor:ConvVar} simply comes from $ \mathrm{Var}(\left\|\nabla F(\widetilde w_\tau)\right\|_2) \leq \EE(\left\|\nabla F(\widetilde w_\tau)\right\|_2^2)\nonumber$. 
\begin{corollary}
\label{cor:ConvVar}
Given the same conditions as that in Theorem~\ref{thm:VRNonCvx} and if we have $\sum_{t=1}^\infty \alpha_t =+\infty$ and $\sum_{t=1}^\infty \alpha_t^2 <+\infty$, then $ \lim_{T\to\infty} \mathrm{Var}\left(\|\nabla F(\widetilde w_\tau)\|_2\right)=0.\nonumber$
\end{corollary}

Finally, we show the convergence result for Online VRADAM. We find that by properly tuning the learning rates, Online VRADAM includes an error which is bounded by $\OO(1/m)$, with the bound with regard to $T$ unchanged.
\begin{theorem}
\label{thm:OnlineVRStrCvx}
Let (1) and (3) in Assumption~\ref{assmpt:LSmoothGrad} be satisfied. Let $\alpha_t^{(k)} = \alpha\gamma^{m-k}/t$ for some $\beta_1 <\gamma<1$, and we require $C_3 = c\beta_1(1-\beta_1)/\sqrt{9G^2 + \epsilon}< 1/\alpha $, where $G$ is given in Lemma~\ref{lem:BoundGradient}. Then for any initial point $w_1$, Online VRADAM with Option (A) satisfies $\EE  F(\widetilde w_T) - F^* \leq \OO\left( T^{-C_3\alpha} + 1/m\right)$.
\end{theorem}

\section{Numerical Experiments}
\subsection{Divergent Example}
\begin{figure}[t]
    \centering
    \begin{subfigure}[t]{0.24\textwidth}
        \includegraphics[width=\textwidth]{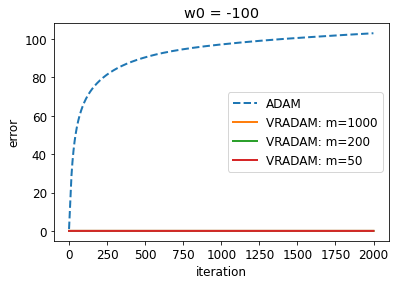}
    \end{subfigure}
    \hfill
    \begin{subfigure}[t]{0.24\textwidth}
        \includegraphics[width=\textwidth]{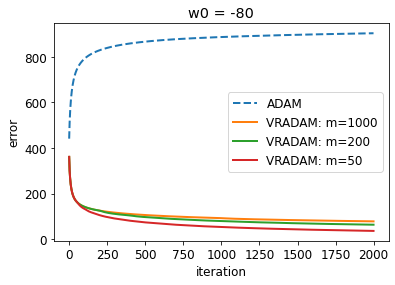}
    \end{subfigure}
    \caption{The error of OP($\delta$) with different initial points}
    \label{fig:div}
\end{figure}

We provide an unconstrained stochastic optimization problem where Adam diverges for any initial point. The goal of this section is to numerically show that Adam diverges on this problem. Since the construction of the mini-batch problem can be equivalently transformed into a stochastic optimization problem with population loss, we stick to the experiments of OP($\delta$) defined in Section 3. Letting $\delta=10$, the optimal solution of OP($\delta$) is $w^*=-\delta^2=-100$. We simulate 1,000 trials for each case and plot the expected L2 error $\EE[(w_t-w^*)^2]$ of $w$. We show in Figure\ref{fig:div} that when $w_0=-100$, Adam diverges while VRADAM keeps the iterate at the optimal point. When $w_0=-80$, Figure~\ref{fig:div} also shows the convergence of VRADAM and the divergence of Adam. This result rules out all of the possible temptations of solving the divergence problem of Adam by using a stochastic initial point.

\subsection{Machine Learning Tasks}
\textbf{Datasets and Implementation:} In this section, we compare the numerical performances of Online VRADAM, VRADAM, and Adam \citep{kingma2014adam} on several real-world classification tasks. The experiments are conducted on the following datasets. \textbf{CovType} \citep{Blackard1998CoverType}: A dataset predicting forest cover type from 54 cartographic variables. The dataset contains 581,012 data points and assigns them into 7 different categories. \textbf{MNIST} \citep{deng2012mnist}: A handwritten digit dataset containing 60k grey level images with size 28$\times$28 pixels. \textbf{NSL-KDD} \citep{tavallaee2009detailed}: A selected subset of the KDD CUP 99 dataset, which is a public dataset used to train a network intrusion detection system. \textbf{Embedded CIFAR-10}: CIFAR-10 \citep{krizhevsky2009learning} consists of 60k color images in 10 classes. We feed each sample to a pretrained ResNet model \citep{he2016deep} and obtain a 1,000 dimensional embedding vector for each image. \textbf{Alpaca:}\citep{alpaca} A dataset of 52,000 instructions and demonstrations generated by OpenAI's text-davinci-003 engine.

We use logistic regression on the four previously mentioned data sets and non-convex deep neural networks on MNIST and Coverage Type datasets. The structures of the deep neural networks used in this section are explained in the appendix.  We also perform parameter finetuning of Llama-7B model \citep{touvron2023llamaopenefficientfoundation} on the Alpaca dataset. Cross entropy is the underlying loss. While training these models, we fix the batch size $|\BB_t|=64$ and the Adam hyper-parameters $\beta_1=0.9,\beta_2=0.999$, which are commonly recommended values in the literature and fine tune the learning rate schedules among $\alpha_t = \alpha_0$, $\alpha_t=\alpha_0/t$ and $\alpha_t = \alpha_0 \gamma ^t$. We perform a grid search among $\alpha_0 \in \{\text{5e-4, 1e-3, 5e-3, 1e-2, 5e-2}\}$ and $\gamma \in\{0.6,0.8,0.95\}$.

In order to eliminate luck from randomness, we ran each experiment with 3 random seeds. We report the average loss of each experiment. We train each setting for 15 epochs for VRADAM and 50 epochs for the other algorithms. While comparing the performances, we display the loss functions up to convergence. The experiments were conducted in PyTorch 1.12.1 on the Google Colab cloud service. We do not apply projections, since in practice they are not needed.

\begin{figure*}[h]
     \centering
     \begin{subfigure}[t]{0.24\textwidth}
         \centering
         \includegraphics[width=\textwidth]{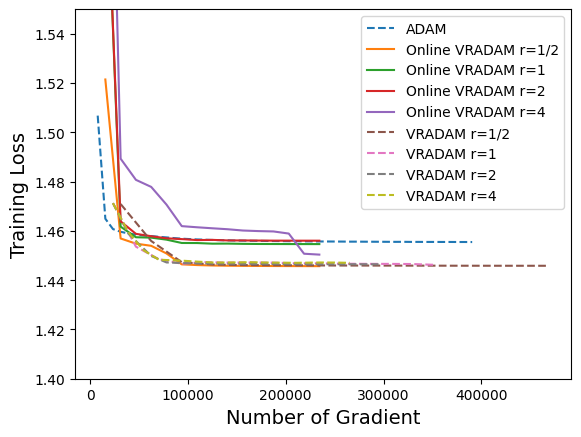}
        \caption{}
         \label{fig:Log_CovType_loss}

     \end{subfigure}
     \hfill
     \begin{subfigure}[t]{0.24\textwidth}
         \centering
         \includegraphics[width=\textwidth]{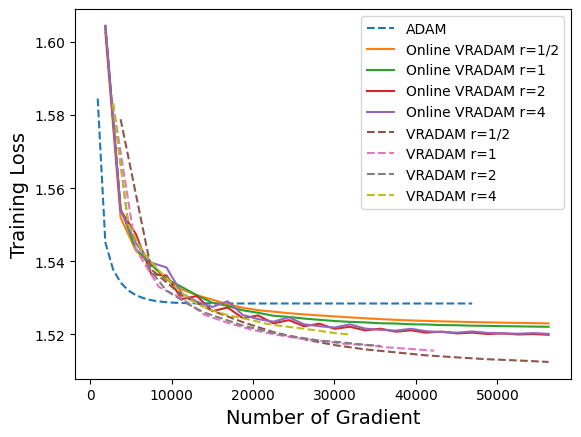}
         \caption{}
         \label{fig:Log_MNIST_loss}
     \end{subfigure}
     \hfill
     \begin{subfigure}[t]{0.24\textwidth}
         \centering
         \includegraphics[width=\textwidth]{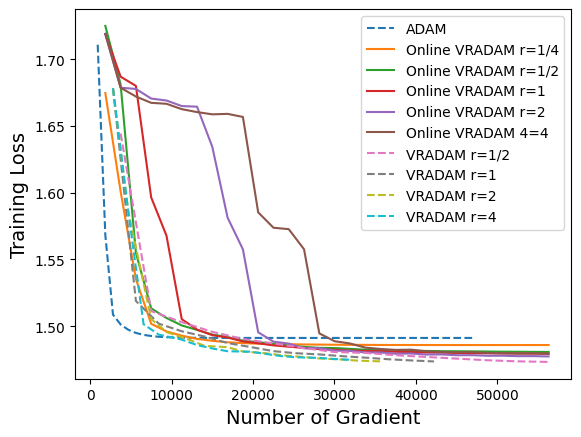}
         \caption{}
         \label{fig:CNN_MNIST_loss}
     \end{subfigure}
     \hfill
     \begin{subfigure}[t]{0.24\textwidth}
         \centering
         \includegraphics[width=\textwidth]{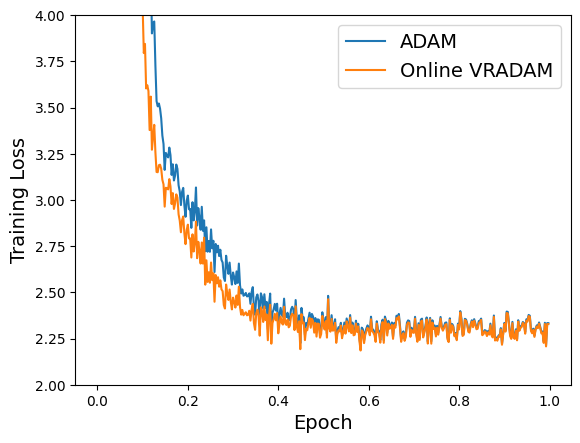}
         \caption{}
         \label{fig:Llama_loss}
     \end{subfigure}
     \vfill
     \begin{subfigure}[t]{0.24\textwidth}
         \centering
         \includegraphics[width=\textwidth]{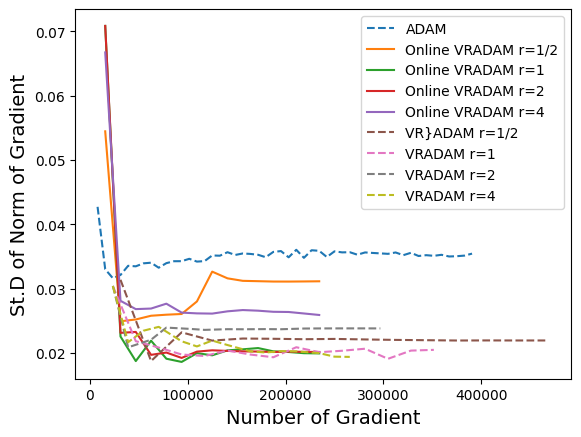}
        \caption{}
         \label{fig:Log_CovType_std}

     \end{subfigure}
     \hfill
     \begin{subfigure}[t]{0.24\textwidth}
         \centering
         \includegraphics[width=\textwidth]{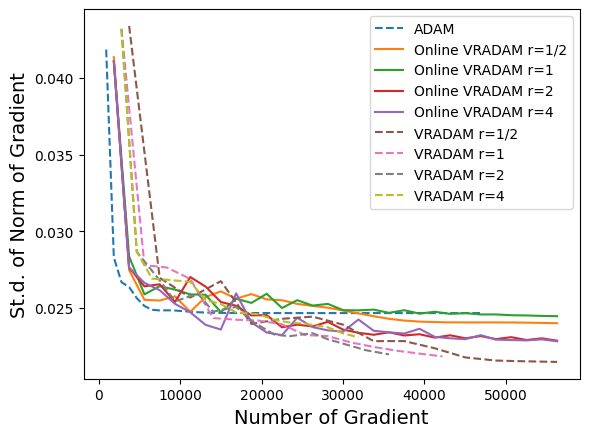}
         \caption{}
         \label{fig:Log_MNIST_std}
     \end{subfigure}
     \hfill
     \begin{subfigure}[t]{0.24\textwidth}
         \centering
         \includegraphics[width=\textwidth]{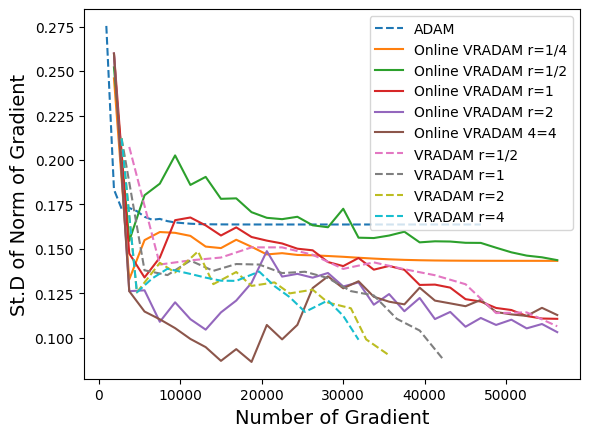}
         \caption{}
         \label{fig:CNN_MNIST_std}
     \end{subfigure}
     \hfill
     \begin{subfigure}[t]{0.24\textwidth}
         \centering
         \includegraphics[width=\textwidth]{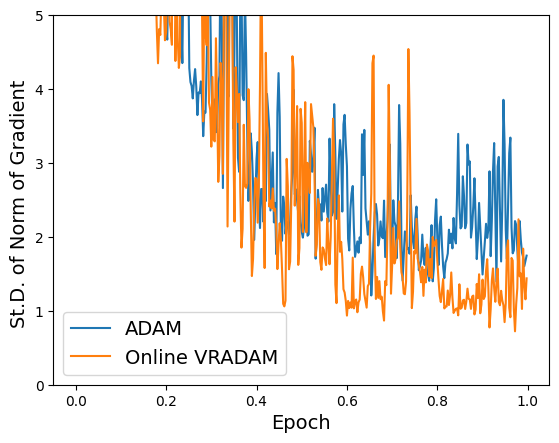}
         \caption{}
         \label{fig:Llama_std}
     \end{subfigure}
      \vfill
     \begin{subfigure}[b]{0.24\textwidth}
         \centering
         \includegraphics[width=\textwidth]{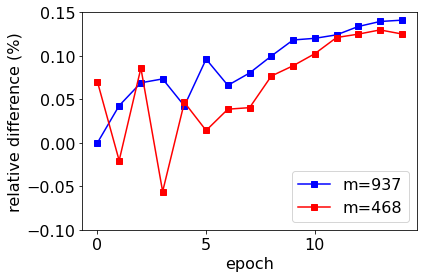}
         \caption{}
         \label{fig:Reset_FFN_MNIST}
     \end{subfigure}
     \begin{subfigure}[b]{0.24\textwidth}
         \centering
         \includegraphics[width=\textwidth]{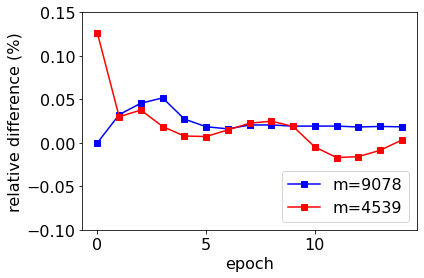}
         \caption{}
         \label{fig:Reset_Log_CovType}
     \end{subfigure}
     \begin{subfigure}[b]{0.24\textwidth}
         \centering
         \includegraphics[width=\textwidth]{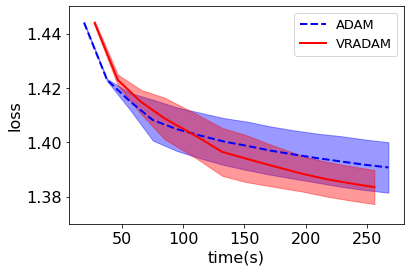}
         \caption{}
         \label{fig:sensitivity_total}
     \end{subfigure}
     \begin{subfigure}[b]{0.24\textwidth}
         \centering
         \includegraphics[width=\textwidth]{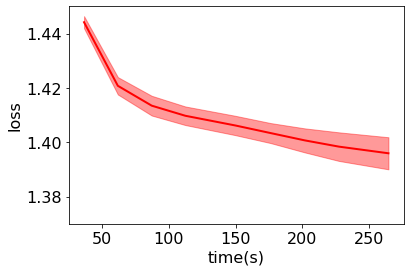}
         \caption{}
         \label{fig:sensitivity_init}
     \end{subfigure}
        \caption{Row 1 and 2: Test losses and standard deviation of norm of gradients for the tasks of  (a,e) Logistic regression on CovType; (b,f) Logistic Regression on MNIST; (c,g) CNN on MNIST; (d, h) finetuning Llama 7b on Alpaca. Row 3: (i,j) Relative differences of loss function of VRADAM without reset over VRADAM with reset on classification tasks of CovType dataset with (i) a feedforward network and (j) logistic regression. (k,l) Deviation of loss. The shaded areas mark the maximal and minimal losses among the three seeds.}
        \label{}
\end{figure*}

\textbf{Main Results:} As for VRADAM, the number of inner loop iterations $m$ in Algorithm~\ref{alg:VRADAM} is also a hyper parameter to decide. If $m$ is too large, the variance reduce procedure does not make a real difference, while an improperly small $m$ would lead to a very frequent computation of the full gradients, which is computationally expensive. We recommend that the length of the inner loop should be about the size of an epoch. In other word, $m \approx N/b$, where $N$ is the number of samples and $b$ is the mini-batch size. In our experiments, we test the performance of VRADAM with $m \in \{N/2b, N/b, 2N/b, 4N/b\}$. It is unfair to compare the convergence of the training loss with regard to the number of epochs. Instead, we compare the training loss relative to the number of gradient computations. Additionally, we evaluate the standard deviation of the gradient norm for each algorithm to demonstrate VRADAM's ability to reduce variance.

\begin{figure}[t]
    \centering
    \includegraphics[width=0.5\linewidth]{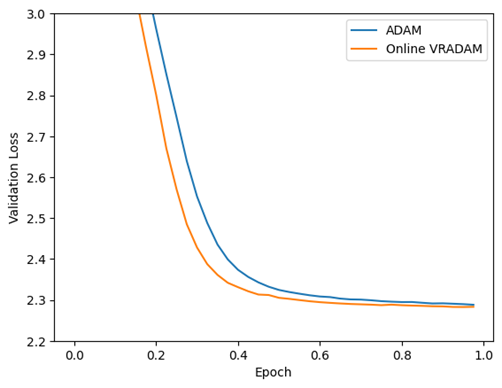}
    \caption{Validation loss of finetuning Llama 7b on Alpaca}
    \label{fig:llm_val}
\end{figure}

Figures~\ref{fig:Log_CovType_loss}-\ref{fig:Llama_loss} illustrate that our variance reduction approach achieves a better convergence rate compared to Adam. Although VRADAM converges slightly slower than Adam in the initial iterations due to the first full gradient computation, it quickly catches up and ultimately surpasses Adam in convergence performance. Furthermore, Figures~\ref{fig:Log_CovType_std}-\ref{fig:Llama_std} show that VRADAM significantly reduces the standard deviation of the gradient norm. In general, the performance of Online VRADAM lies between that of the original VRADAM and Adam. Notably, Online VRADAM exhibits a faster convergence rate than Adam on the LLM task. We recommend using Online VRADAM in scenarios with limited computational resources or when rapid convergence is required. In order to evaluate the validation performance of VRADAM on the non-LLM tasks, we report in Table~\ref{tab:test_acc} the validation accuracies of VRADAM using the optimal choice $r = mb/N$, together with those of Adam. In addition, we further assess VRADAM on the LLM task by examining the validation loss. As shown in Figure~\ref{fig:llm_val}, Online VRADAM achieves lower validation loss than ADAM, indicating improved generalization performance in LLM training.

As mentioned previously, we recommend the resetting option. Figures~\ref{fig:Reset_FFN_MNIST} and~\ref{fig:Reset_Log_CovType} display the performance of the resetting option in a few experiments and show that the resetting option helps with the convergence of VRADAM. Additional experiments are in the appendix.

\begin{table}
    \centering
    \begin{tabular}{ccc}
     \hline   Experiment  &  VRADAM (optimal $r$)  &  Adam\\
    \hline Logistic CovType&72.0 & 71.4\\
     FFN CovType&80.5& 79.0\\
     Logistic MNIST& 93.3 & 93.3\\
     Logistic NSL-KDD & 99.9 & 99.9\\\hline
    \end{tabular}
    \caption{Validation Accuracies for VRADAM with the optimal $r=mb/N$ values and Adam (\%)}
    \label{tab:test_acc}
\end{table}

In conclusion, we observe that VRADAM outperforms Adam on large datasets, such as CovType. Such datasets may contain extreme values, which may harm the convergence of Adam. We find that the computational cost when calculating the full gradients can be compensated by the benefits of quick convergence of VRADAM. We recommend VRADAM over Adam for tasks with large datasets, in particular if loss is convex. We remark that the limitation of VRADAM is the computation of full gradients, as it may be costly for some tasks, such as training LLMs, which can be solved by our approach of Online VRADAM.

\textbf{Sensitivity:} We consider the CovType classification task with the FFN model as an example to analyze the sensitivity of our algorithm. As stated previously, our algorithm fixes the convergence issue of Adam by reducing the variance. Our experiments take three different seeds and the deviation of the results reflects the variance of the algorithm.  Figure~\ref{fig:sensitivity_total} shows that VRADAM reduces the noise comparing with Adam. We also studied the sensitivity of VRADAM over the different initial points, as shown in Figure~\ref{fig:sensitivity_init}.

\newpage

% \section*{Impact Statement}
% This paper presents work whose goal is to advance the field of Machine
% Learning. There are many potential societal consequences of our work, none
% which we feel must be specifically highlighted here.
\bibliography{example_paper}
\bibliographystyle{icml2026}

%%%%%%%%%%%%%%%%%%%%%%%%%%%%%%%%%%%%%%%%%%%%%%%%%%%%%%%%%%%%%%%%%%%%%%%%%%%%%%%
%%%%%%%%%%%%%%%%%%%%%%%%%%%%%%%%%%%%%%%%%%%%%%%%%%%%%%%%%%%%%%%%%%%%%%%%%%%%%%%
% APPENDIX
%%%%%%%%%%%%%%%%%%%%%%%%%%%%%%%%%%%%%%%%%%%%%%%%%%%%%%%%%%%%%%%%%%%%%%%%%%%%%%%
%%%%%%%%%%%%%%%%%%%%%%%%%%%%%%%%%%%%%%%%%%%%%%%%%%%%%%%%%%%%%%%%%%%%%%%%%%%%%%%
\newpage
\appendix
\onecolumn

\section{Resetting/ No Resetting Options}
We provide two options with regard to the update of Adam states. In Option (A), we reinitialize the Adam states at the beginning of each outer iteration, while Option (B) keeps the state through the whole training process. Although for the original Adam, resetting the states harms the performance of the algorithm, we computationally found that the resetting option works better in VRADAM. Intuitively, this is because in each inner loop, the first step $g_1^{(t)}$ is always the full gradient direction, which makes a more efficient update than the direction adapted by previous Adam states. In order to support our argument, we provide a theoretical analysis of an example. If we fix the initial point $w_1\in\RR$ and the mini-batch losses $F^{\BB_1^{(1)}}, F^{\BB_2^{(1)}},\ldots, F^{\BB_m^{(1)}}$, the iterates are identical between the two options through the $t=1$ iteration. We consider the objective values after the first update in the second outer iteration, i.e. $F\left(w_2^{(2)}\right)$. 
 We make the following assumptions.
 
\begin{assumption}\label{assmpt:Noreset} The framework described in this section satisfies: \textbf{(1)} $F(w)$ is $c-$strongly convex, and its gradient is $L-smooth$. Each one of $\left|g_1^{(1)}\right|,\ldots,\left|g_m^{(1)}\right|$ and $\left|g_1^{(2)}\right|$ is upper bounded by $G>0$. \textbf{(2)} The algorithm makes progress in the $t=1$ iteration, specifically, $\left|m^{(1)}_{m+1}\right| \geq \left|F'\left(w^{(2)}_{1}\right)\right|$. \textbf{(3)} The hyper-parameters satisfy $L\alpha_2\geq 2\sqrt{G^2 + \epsilon}$ and $ \frac{L}{c} \leq \frac{2\beta_1-1}{1-\beta_1^{m+1}}\sqrt{\frac{\epsilon}{G^2 + \epsilon}}.$
\end{assumption}

Notice that the second assumption assumes that the exponential moving average of the steps in the first loop is larger than the full gradient at the beginning of the second loop, which reflects that the algorithm makes progress in the first iteration. The third assumption can be satisfied by $\beta_1$ that is close enough to $1$ and appropriately selected $\alpha_2$. Our concern is to compare the objective values $F\left(w^{(2)}_2 \right)$ and $F\left( \hat w^{(2)}_2\right)$ of options A and B respectively. The following theorem shows that option A, i.e., the option where the states of Adam are reset at the beginning of each outer iteration, makes more efficient descent, hence works better. While Theorem~\ref{thm:ResetBetter} allows the preference of option A within the two outer iterations, the computational experiments confirm this choice in general.
\begin{theorem}
\label{thm:ResetBetter}
Given Assumption~\ref{assmpt:Noreset}, $F(\hat w^{(2)}_2) \geq F(w^{(2)}_2 )$.
\end{theorem}
The proof is given in the following section.

\section{Proofs}
\subsection{Technical Lemmas}
\begin{lemma}
\label{lem:Jensen}
In a probability space $(\Omega, \FF, \PP)$, let there be an $\FF$-measurable random variable $X(w)$ and an event $A\in\FF$ such that $\PP\{A\}>0$. For a convex function $\phi(x)$, we have
$$\frac{\EE \left[ \phi(X)\mathbf{1}\{A\}\right]}{\PP\{A\}}\geq  \phi\left( \frac{\EE\left[X \mathbf{1}\{A\}\right]}{\PP\{A\}} \right).$$
\end{lemma}

\textbf{Proof:}
Let 
$$x_0 = \frac{\EE\left[X\mathbf{1}\{A\}\right]}{\PP\{A\}} = \frac{1}{\PP\{A\}}\int_A X(w) d\PP(w).$$

Since $\phi$ is convex, there exists a sub-gradient of $\phi$ at $x_0$, i.e., there exists an $a$ such that
$$\phi(x) \geq \phi(x_0) + a(x-x_0)$$
for any $x\in\RR$. Then we have
\begin{eqnarray}
\frac{\EE\left[\phi(X) \mathbf{1}\{A\}\right]}{\PP\{A\}} &=& \frac{1}{\PP\{A\}}\int_A \phi(X(w))d\PP(w) \nonumber\\
&\geq& \frac{1}{\PP\{A\}}\int_A a (X(w) - x_0) + \phi(x_0) d\PP(w) \nonumber\\
&=& \frac{a}{\PP\{A\}} \int_A X(w) d\PP(w) + \frac{\phi(x_0) - ax_0}{\PP\{A\}} \int_A 1d\PP(w)\nonumber\\
&=& ax_0 +\phi(x_0) - ax_0\nonumber \\
&=& \phi(x_0)\nonumber,
\end{eqnarray}
which finishes the proof.

\begin{lemma}
\label{lem:simplecalculus}
For any $x,y >0$, 
$$(x+y)^{3} \leq 4(x^3 + y^3).$$
\end{lemma}

\textbf{Proof}:
For $t\geq 0$, let
\begin{eqnarray}
h(t):=\frac{(1+t)^3}{1+t^3} = 1 + 3 \frac{t + t^2}{1 + t^3}.\nonumber
\end{eqnarray}

The derivative of $h(t)$ reads
\begin{eqnarray}
h'(t) &=& 3\frac{(1+2t)(1+t^3) - 3t^2(t+t^2)}{(1+t^3)^2} = -3\frac{(t-1)(t+1)^3}{(t^3+1)^2}.\nonumber
\end{eqnarray}

Apparently $h(t)$ achieves the maximum at $t=1$, thus $h(x/y)\leq h(1)=4$, and we have
\begin{equation*}
    \frac{(x+y)^3}{x^3 + y^3}\leq 4.
\end{equation*}

\begin{lemma}
\label{lem:sumbetaalpha}
Given $\alpha_t = \alpha/t$ and $\beta_1\in[0,1)$, there exists a constant $\bar C>0$, such that for any $t\geq 2$ we have 
$$\sum_{j=1}^{t-1}\alpha_j \beta_1^{t-j}\leq \bar C\alpha_t .$$
\end{lemma}

\textbf{Proof:} Letting $t^* =\left \lfloor\frac{t-1}{2}\right\rfloor$, we have
\begin{eqnarray}
\sum_{j=1}^{t-1} \alpha_j \beta_1^{t-j} &=& \sum_{j=1}^{t^*} \alpha_j \beta_1^{t-j} + \sum_{j=t^*+1}^{t-1} \alpha_j \beta_1^{t-j} \nonumber\\
&\leq& \alpha \sum_{j=1}^{t^*} \beta_1^{t-j} + \frac{\alpha}{t^* + 1} \sum_{j=t^*+1}^{t-1} \beta_1^{t-j} \nonumber \\
&\leq&  \frac{\alpha\beta_1^{t-t^*}}{1-\beta_1} + \frac{\alpha}{t^* + 1} \frac{\beta_1}{1-\beta_1} \nonumber\\
&\leq& \frac{\alpha \beta_1^{(t+1)/2}}{1-\beta_1} + \frac{2\alpha\beta_1}{(1-\beta_1)} \frac{1}{t-1} = \OO(t^{-1}).\nonumber
\end{eqnarray}
Thus there exists a positive constant $\bar C$ such that for any $t\geq 2$, 
$$\sum_{j=1}^{t-1} \alpha_j\beta_1^{t-j} \leq \bar C \alpha_t.$$

\begin{lemma}
\label{lem:Recursive}
Consider $0<A<1$ and $T\geq 2$, and let 
\begin{eqnarray}
\lambda_{T-1} &=& \prod_{t=1}^{T-1} \left(1-\frac{A}{t}\right),\nonumber\\
\nu_{T-1} &=& \sum_{t=1}^{T-1} \frac{1}{t^2} \prod_{j=t+1}^{T-1}\left(1-\frac{A}{j}\right).\nonumber
\end{eqnarray}
Then we have
\begin{eqnarray}
\lambda_{T-1} \leq \OO(T^{-A})\nonumber
\end{eqnarray}
and
\begin{eqnarray}
\nu_{T-1} \leq \OO(T^{-A}).\nonumber
\end{eqnarray}
\end{lemma}
\textbf{Proof:} Notice that
\begin{eqnarray}
\log \lambda_{T-1} = \sum_{t=1}^{T-1} \log\left(1-\frac{A}{t}\right)\leq -A\sum_{t=1}^{T-1}\frac{1}{t}\leq -A\log T,\nonumber
\end{eqnarray}
where the first inequality comes from $\log(1-x)\leq-x$ for $x\geq0$ and the second inequality uses the integral approximation
$$\sum_{t=1}^{T-1}\frac{1}{t} \geq \sum_{t=1}^{T-1}\int_{t}^{t+1} \frac{1}{s}\;ds =\int_{1}^T \frac{1}{s}\;ds=\log T.$$
Thus $
\lambda_{T-1} \leq T^{-A}. $
Similarly, we have
\begin{eqnarray}
\log \frac{\lambda_{T-1}}{\lambda_t} = \sum_{k={t+1}}^{T-1} \log\left( 1- \frac{A}{t}\right)\leq - A\log\frac{T}{t+1},\label{eq:boundprod}
\end{eqnarray}
and then,
\begin{eqnarray}
\nu_{T-1} \leq \sum_{t=1}^{T-1} \frac{1}{t^2} \left(\frac{T}{t+1}\right)^{-A} \leq  T^{-A}\sum_{t=1}^{T-1} \frac{(t+1)^{A}}{t^2}\leq  2^{A} T^{-A} \sum_{t=1}^{T-1} t^{-2+A} .\nonumber
\end{eqnarray}
Again, applying the integral approximation yields
\begin{eqnarray}
\sum_{t=1}^{T-1} t^{-2+A} \leq 1 + \sum_{t=2}^{T-1}\int_{t-1}^t s^{-2+A}\;ds = 1+\int_{1}^{T-1} s^{-2+A}\;ds\leq 1 + \frac{1}{1-A} = \frac{2-A}{1-A}<\infty.\nonumber
\end{eqnarray}
Then we have $\nu_{T-1} \leq \OO( T^{-A}).$

\begin{lemma}
\label{lem:StrCvxDiff}
    Given a $c$-strongly convex function $f:\RR^d \to \RR$ with global minimum $x^*$ and $L$-smooth gradient, we have
    \begin{enumerate}
        \item $f\left(x^* + t_2 a\right) \geq f\left(x^* + t_1 a\right) + \frac{c}{2}\left(t_2^2-t_1^2\right)$, for all $0<t_1<t_2$ and $a\in\RR^d$ such that $\left\| a\right\|_2 = 1$, and 
        \item $f\left(x^* + ta_2\right) \geq f\left(x^*+ta_1\right) -\frac{L^2}{2c}t^2$, for all $t>0$ and $a_1,a_2\in\RR^d$ such that $\left\|a_1\right\|_2 = 1$.
    \end{enumerate}
\end{lemma}
\textbf{Proof:} (i) Letting $g(t) = f(x^*+ta)$, its derivative is $g'(t)=\nabla f(x^*+ta)^\top a$. For any $t$ and $t'$, we have
\begin{eqnarray}
    (g'(t) - g'(t'))(t-t') &=& \left(\nabla f(x^* + ta) - \nabla f(x^* + t'a)\right)^\top (t-t')a\nonumber\\
    &=& \left(\nabla f(x^* + ta) - \nabla f(x^* + t'a)\right)^\top (x^*+ta-x^*-t'a)\nonumber\\
    &\geq& c\left\|(t-t')a\right\|_2^2 \nonumber\\
    &=& c(t-t')^2.\nonumber
\end{eqnarray}
The inequality holds because $f$ is strongly convex. This implies that $g(t)$ is a $c$-strongly convex function. Furthermore, by letting $t'=0$ and $t>0$, we have 
\begin{eqnarray}
    g'(t) \geq ct.\nonumber
\end{eqnarray}
Then for any $0<t_1<t_2$, we obtain
\begin{eqnarray}
    f(x^*+t_2a) - f(x^*+t_1a) &=& g(t_2) - g(t_1)\nonumber\\
    &\geq& g'(t_1) (t_2 - t_1) + \frac{c}{2}(t_2-t_1)^2\nonumber\\
    &\geq& ct_1(t_2-t_1) + \frac{c}{2}(t_2-t_1)^2 = \frac{c}{2}(t_2^2 - t_1^2),\nonumber
\end{eqnarray}
where the first inequality holds because $g(t)$ is strongly convex.

(ii) Using strong convexity of $f$, we obtain
\begin{eqnarray}
    f(x^*+ta_2) &\geq& f(x^*+ta_1) + t\nabla f(x^*+ta_1)^\top(a_2-a_1) + \frac{c}{2}\left\|t(a_2-a_1)\right\|_2^2 \nonumber\\
    &\geq& f(x^*+ta_1) - t\left\|\nabla f(x^*+ta_1)\right\|_2\left\|a_2-a_1\right\|_2 + \frac{ct^2}{2}\left\|a_2-a_1\right\|_2^2 \nonumber\\
    &\geq& f(x^*+ta_1) - Lt^2\left\|a_2-a_1\right\|_2 + \frac{ct^2}{2}\left\|a_2-a_1\right\|_2^2,\nonumber
\end{eqnarray}
where the last inequality holds because $\nabla f(x)$ is Lipschitz smooth and thus,
\begin{eqnarray}
    \left\|\nabla f(x^*+ta_1)\right\|_2&=&\left\|\nabla f(x^*+ta_1) - \nabla f(x^*)\right\|_2 \nonumber\\
    &\leq& L\left\|x^* + ta_1 - x^* \right\|_2 = Lt.\nonumber
\end{eqnarray}
Thus, we have
\begin{eqnarray}
    f(x^*+ta_2) &\geq& f(x^*+ta_1) + \frac{ct^2}{2}\left(\left\|a_2-a_1\right\|_2^2 - \frac{2L}{c}\left\|a_2-a_1\right\|_2 + \frac{L^2}{c^2}\right) - \frac{L^2t^2}{2c}\nonumber\\
    &=& f(x^*+ta_1) + \frac{ct^2}{2}\left(\left\|a_2-a_1\right\|_2- \frac{L}{c}\right) ^2 - \frac{L^2t^2}{2c}\nonumber\\
    &\geq& f(x^*+ta_1) - \frac{L^2t^2}{2c},\nonumber
\end{eqnarray}
which completes the proof.

\color{black}

\subsection{Proof of Theorem~\ref{thm:DivOP}}

Let $p = \PP(\xi=1)$. In each step, the update value is
\begin{eqnarray}
\Delta_t &=& -\frac{\alpha_t g_t}{\sqrt{\beta_2v_{t-1} + (1-\beta_2)g_t^2 }}\nonumber\\
&=&\begin{cases}
-\frac{\alpha_t(w_t/\delta + \delta^4) }{\sqrt{\beta_2 v_{t-1} + (1-\beta_2) (w_t/\delta + \delta^4)^2 }} & \text{with probability }\; p \\
\frac{\alpha_t(1-w_t/\delta)}{\sqrt{\beta_2 v_{t-1} + (1-\beta_2)(1-w_t/\delta)^2}} & \text{with probability }\; 1-p.
\end{cases}\nonumber
\end{eqnarray}
Apparently, 
\begin{equation*}
F(w) =  \frac{w^2}{2\delta} + \delta w,
\end{equation*}
and 
\begin{equation*}
w^*=-\delta^2.
\end{equation*}

We use contradiction to prove the theorem. Assume that $\EE[F(w_t)-F(w^*)]\to 0$. Notice that
\begin{equation*}
F(w_t) - F(w^*) = \frac{1}{2\delta} (w_t-w^*)^2\nonumber,
\end{equation*}
which means that $\EE[F(w_t)-F(w^*)]\to 0$ is equivalent to  $\EE\left[\left(w_t-w^*\right)^2\right]\to 0$. Let us select $0 < \epsilon < 1/2$, and we choose $T_\epsilon$ such that $t>T_\epsilon$ implies $\EE\left[\left(w_t-w^*\right)^2\right]<\epsilon$. The following discussion is based on $w_t$ such that $t>T_\epsilon$.

We have\begin{equation}
\label{stplth}
|\Delta_t|= \frac{\alpha_t |g_t|}{\sqrt{\beta_2 v_{t-1} +(1-\beta_2) g_t^2}} \leq\frac{\alpha_t |g_t|}{\sqrt{(1-\beta_2) g_t^2}} = \frac{\alpha_t}{\sqrt{1-\beta_2}},
\end{equation}
where the inequality is due to $v_k$ being non-negative for any $k$.

Let $\FF_t$ be the filtration including all the information obtained until the update of $w_t$, including $w_t$. We define the following event
\begin{eqnarray}
E := \left\{|w_t - w^*|< \delta^2 \right\}\nonumber
\end{eqnarray}
which is known given $\FF_t$. We have
$$\PP\{E^c\} = \PP\left\{ |w_t-w^*|\geq\delta^2\right\} \leq\frac{\EE\left[(w_t-w^*)^2\right]}{\delta^4} < \frac{\epsilon}{\delta^4}.$$
Given $E^c$, we simply bound the step size with the lower bound
\begin{eqnarray}
\EE\left[ \Delta_t \mathbf{1}\{E^c\}\right] \geq -\frac{\alpha_t}{\sqrt{1-\beta_2}} \EE\left[ \mathbf{1}\{E^c\}\right] \geq -\frac{\epsilon}{\delta^4}\frac{\alpha_t}{\sqrt{1-\beta_2}}.\nonumber
\end{eqnarray}
For the samples in $E$, we have
\begin{eqnarray}
\EE\left[ \Delta_t \mathbf{1}\{E\}\right] &=& \EE\left[\EE\left[\Delta_t \mathbf{1}\{E\} \left|\FF_t\right.\right]\right] 
=  \EE\left[\EE\left[\Delta_t  \left|\FF_t\right.\right]\mathbf{1}\{E\}\right] \nonumber\\
&=& \EE\left[ \mathbf{1}\{E\} \left\{ (1-p) \frac{\alpha_t(1-w_t/\delta) }{\sqrt{\beta_2 v_{t-1} + (1-\beta_2) (1-w_t/\delta)^2}}\right\}\right]\nonumber\\
&&- \EE\left[ \left\{ p \frac{\alpha_t(w_t/\delta + \delta^4)}{\sqrt{\beta_2 v_{t-1} +(1-\beta_2) (w_t/\delta+\delta^4)^2}}\right\}\right]\nonumber\\
&\geq &\EE \left[\mathbf{1}\{E\} (1-p)\frac{\alpha_t}{\sqrt{\beta_2 v_{t-1} + (1-\beta_2)(1+2\delta)^2}}\right] \nonumber\\
&&- p \frac{\alpha_t}{\sqrt{1-\beta_2}}\PP\{E\} .\label{Eq3}
\end{eqnarray}
In the inequality, the first term is bounded because $-2\delta<w_t/\delta<0$ by the definition of $E$ and the second term is bounded by the bound of the step length in (\ref{stplth}). 

By applying Lemma~\ref{lem:Jensen} to (\ref{Eq3}), we have
\begin{eqnarray}
\EE[\Delta_t\mathbf{1}\{E\}] \geq (1-p)\PP(E)\frac{\alpha_t}{\sqrt{\beta_2 \EE[v_{t-1} \mathbf{1}\{E\}]/\PP\{E\} +(1-\beta_2)(1+2\delta)^2 }} - p\PP(E)\frac{\alpha_t}{\sqrt{1-\beta_2}}.\nonumber
\end{eqnarray}
We next focus on the conditional expectation
\begin{eqnarray}
\EE\left[v_{t-1}\mathbf{1}\{E\}\right]
&=&  (1-\beta_2)\sum_{k=1}^{t-1}\beta_2^{t-1-k}\EE\left[\mathbf{1}\{E\} g_k^2\right]. \nonumber
\end{eqnarray}
We claim that for any trajectory in $E$ and for any $k<t$, we have
$$\left|w_t-w_k\right| = \left|\sum_{j=k}^{t-1}\Delta_j \right| \leq \sum_{j=k}^{t-1}\left|\Delta_j\right| \leq \frac{\alpha(t-k)}{\sqrt{1-\beta_2}}.  $$
The last inequality comes from the bound of the step length in (\ref{stplth}). Then we have
$$ w_t - \frac{\alpha(t-k)}{\sqrt{1-\beta_2}}\leq w_k \leq w_t + \frac{\alpha(t-k)}{\sqrt{1-\beta_2}}.$$
Let us recall that given $E$, we have $-2\delta ^2 < w_k < 0$, and therefore
\begin{equation}
\label{wkbd}
    -2\delta^2 - \frac{\alpha(t-k)}{\sqrt{1-\beta_2}}\leq w_k \leq  \frac{\alpha(t-k)}{\sqrt{1-\beta_2}}.
\end{equation} 
Then for each $k=1,\ldots,t-1$, we obtain
\begin{eqnarray}
\EE \left[g_k ^2 \mathbf{1}\{E\}\right] &=&  \EE\left[\EE[g_k^2\mathbf{1}\{E\} \left|\FF_k\right.]\right]\nonumber\\
&\leq& \EE\left[\left(\EE\left[ g_k^{2(1+\mu)}|\FF_k \right]\right)^{1/(1+\mu)}\left(\EE \left[ \mathbf{1}\{E\}\right]\right)^{\mu/(1+\mu)} \right]\nonumber\\
&\leq& \EE\left[\left(\EE\left[ g_k^{2(1+\mu)}|\FF_k \right]\right)^{1/(1+\mu)} \right]\nonumber
\end{eqnarray}
where the inequality holds for any $\mu$ according to the Holder inequality. Let $0<\mu<1/2$. According to the bound given previously in (\ref{wkbd}) and $\delta \geq 2$, we have
\begin{eqnarray}
\left( \frac{w_k}{\delta} + \delta^4 \right)^2 &\leq& \left( \frac{\alpha(t-k)}{\delta \sqrt{1-\beta_2}} + 2 \delta + \delta^4\right)^2\nonumber\\
\left( 1- \frac{w_k}{\delta}\right)^2 &\leq& \left( \frac{\alpha(t-k)}{\delta \sqrt{1-\beta_2}} + 2\delta + 1\right)^2.\nonumber
\end{eqnarray} 
Then we derive, 
\begin{eqnarray}
\EE\left[ g_k^{2(1+\mu)}|\FF_k \right] &=& p \left( \frac{w_k}{\delta} + \delta ^4 \right)^{2(1+\mu)} + (1-p) \left( \frac{w_k}{\delta} -1\right)^{2(1+\mu)}\nonumber \\
&\leq& \frac{1+\delta}{\delta ^4 }\left( \frac{\alpha(t-k)}{\delta \sqrt{1-\beta_2}} + 2\delta + \delta ^ 4\right)^{2(1+\mu)} + \left( \frac{\alpha(t-k)}{\delta \sqrt{1-\beta_2}} + 2\delta + 1\right)^{2(1+\mu)} \nonumber\\
&=& (1+\delta) \left(\frac{\alpha(t-k)}{\delta^5 \sqrt{1-\beta_2}} +\frac{2}{\delta ^ 3}+1\right)^{2(1+\mu)} \delta^{4+8\mu}+\left( \frac{\alpha(t-k)}{\delta \sqrt{1-\beta_2}} + 2\delta + 1\right)^{2(1+\mu)}\nonumber\\
&\leq& 2\delta \cdot \left(\frac{\alpha(t-k)}{\delta^5 \sqrt{1-\beta_2}} +\frac{2}{\delta ^ 3}+1\right)^{3}\cdot \delta ^{4 + 8\mu}+\left( \frac{\alpha(t-k)}{\delta \sqrt{1-\beta_2}} + 2\delta + 1\right)^3\nonumber \\
&\leq& 2\left(\frac{\alpha(t-k)}{\sqrt{1-\beta_2}} + 2\right)^3 \delta ^{5+8\mu} + \left( \frac{\alpha(t-k)}{\sqrt{1-\beta_2}} +3\delta\right)^3  \nonumber \\
&\leq& \left(\frac{8\alpha^3 (t-k)^3}{(1-\beta_2)^{3/2}} + 64\right) \delta ^{5+8\mu} + \frac{4\alpha^3 (t-k)^3}{(1-\beta_2)^{3/2}} + 108 \delta^3.\nonumber
\end{eqnarray}
We have used $\delta>2$ and $0<\mu<1/2$. The last inequality holds because of Lemma~\ref{lem:simplecalculus}. Then we obtain
\begin{eqnarray}
\left(\EE\left[ g_k^{2(1+\mu)}|\FF_k \right]\right)^{1/(1+\mu)} &\leq& \left[\left(\frac{8\alpha^3 (t-k)^3}{(1-\beta_2)^{3/2}} + 64\right) \delta ^{5+8\mu} + \frac{4\alpha^3 (t-k)^3}{(1-\beta_2)^{3/2}} + 108 \delta^3\right]^{1/(1+\mu)}\nonumber \\
&=& \left(\frac{8\alpha^3 (t-k)^3}{(1-\beta_2)^{3/2}} + 64  + \frac{4\alpha^3 (t-k)^3}{(1-\beta_2)^{3/2}}\delta ^{-5-8\mu} + 108 \delta^{-2-8\mu}\right)^{1/(1+\mu)} \delta ^{(5 + 8\mu)/(1+\mu)}\nonumber \\
&\leq& \left( \frac{12 \alpha ^3(t-k)^3}{(1-\beta_2)^{3/2}} + 172\right)^{1/(1+\mu)} \delta ^{(5+8\mu)/(1+\mu)} \nonumber\\
&\leq& \left( \frac{12 \alpha ^3(t-k)^3}{(1-\beta_2)^{3/2}} + 172\right) \delta ^{(5+8\mu)/(1+\mu)}.\nonumber
\end{eqnarray}
The third inequality uses $\delta>1$ and thus
\begin{eqnarray}
\EE[v_{t-1}\mathbf{1}\{E\}] &\leq&  (1-\beta_2)\sum_{k=1}^{t-1}\beta_2^{t-1-k}\left( \frac{12 \alpha ^3(t-k)^3}{(1-\beta_2)^{3/2}} + 172\right) \delta ^{(5+8\mu)/(1+\mu)} \nonumber \\
&\leq& \delta^{(5+8\mu)/(1+\mu)} \left\{ \frac{12\alpha^3}{\sqrt{1-\beta_2}}\sum_{k=1}^\infty \beta_2^{k-1}k^3 + 172 \right\}\nonumber\\
&\leq& \delta^{(5+8\mu)/(1+\mu)} \left\{ \frac{72\alpha^3}{(1-\beta_2)^{9/2}} + 172 \right\}:= M_1 \delta^{(5+8\mu)/(1+\mu)}\nonumber
\end{eqnarray}
where the last inequality is because $\sum_{k=1}^\infty \beta_2^{k-1}k^3 = (1+4\beta_2+\beta_2^2)/(1-\beta_2)^4<6/(1-\beta_2)^4$. Thus, we have
\begin{eqnarray}
\EE\left[\Delta_t\mathbf{1}\{E\}\right]&\geq& \PP\{E\}\left\{ \left(1-\frac{1+\delta}{1+\delta^4}\right)\frac{\alpha_t}{\sqrt{\beta_2M_1 \delta ^{(5+8\mu)/(1+\mu)}/\PP\{E\}+ (1-\beta_2)(1+2\delta)^2}}\right.\nonumber\\
&&\left.- \frac{1+\delta}{1+\delta^4}\frac{\alpha_t}{\sqrt{1-\beta_2}}\right\}\nonumber\\
&\geq& \frac{1}{2}\left\{ \left(1-\frac{1+\delta}{1+\delta^4}\right)\frac{\alpha_t}{\sqrt{2\beta_2M_1 \delta ^{(5+8\mu)/(1+\mu)}+ (1-\beta_2)(1+2\delta)^2}}\right.\nonumber\\
&&\left.- \frac{1+\delta}{1+\delta^4}\frac{\alpha_t}{\sqrt{1-\beta_2}}\right\},\nonumber
\end{eqnarray}
where the second inequality follows from
$$\PP\{E\} >1-\frac{\epsilon}{\delta^4} >1 - \epsilon > \frac{1}{2}.$$
Then the full expectation of $\Delta_t$ is
\begin{eqnarray}
\EE\left[\Delta_t\right]&\geq& \frac{1}{2}\left\{ \left(1-\frac{1+\delta}{1+\delta^4}\right)\underbrace{\frac{\alpha_t}{\sqrt{2\beta_2 M_1 \delta ^{(5+8\mu)/(1+\mu)}+ (1-\beta_2)(1+2\delta)^2}}}_{T_1}\right.\nonumber\\
&&\left.- \underbrace{\frac{1+\delta}{1+\delta^4}\frac{\alpha_t}{\sqrt{1-\beta_2}}}_{T_2}\right\} - \underbrace{\frac{1}{2\delta^4} \frac{\alpha_t}{\sqrt{1-\beta_2}}}_{T_3}.\nonumber
\end{eqnarray}
Notice that $T_1 =\Omega(\delta^{-(5+8\mu)/(2+2\mu)})$, $T_2=\OO(\delta^{-3})$ and $T_3=\OO(\delta^{-4})$. As long as we set $\mu<1/2$, we have $(5+8\mu)/(2+2\mu) < 3 < 4$, thus the right hand side can be positive for sufficiently large $\delta$, only dependent on $\beta_2$. We conclude that we can assume $\EE[\Delta_t]>c_0 \alpha_t >0$. This means $w_t$ keeps drifting in the positive direction. Then for any $k\geq 1$ and $t>T_\epsilon$, we have
\begin{eqnarray}
\EE\left[(w_{t+k}-w^*)^2\right] &=& \EE\left[(w_{t+k}-w_t)^2\right] + 2\EE\left[(w_{t+k}-w_t)(w_t-w^*)\right] + \EE\left[(w_t-w^*)^2\right]\nonumber \\
&\geq& \EE\left[(w_{t+k}-w_t)^2\right] - 2 \sqrt{\EE\left[(w_{t+k}-w_t)^2  \right]\EE\left[(w_t-w^*)^2\right]} + \EE\left[ (w_t-w^*)^2\right]\nonumber\\
&=& \left(\sqrt{\EE\left[(w_{t+k}-w_t)^2\right]} - \sqrt{\EE\left[(w_t-w^*)^2\right]}\right)^2 ,\label{Thm1Last}
\end{eqnarray}
where the inequality is the Cauchy-Schwartz inequality for random variables. According to the assumption that $\sum_{t=1}^\infty\alpha_t = \infty$, there exists $k$ large enough such that $\sum_{l=0}^{k-1} \alpha_{t+l} > 3\sqrt{\epsilon}/{c_0}$, which implies that $\EE[(w_{t+k} - w_t)^2]\geq (\EE[w_{t+k}-w_t])^2\geq c_0^2(\sum_{l=1}^{k-1} \alpha_{t+l})^2 \geq 9\epsilon$, and thus from (\ref{Thm1Last}) we have
\begin{equation*}
\EE\left[(w_{t+k}-w^*)^2\right] \geq (c_0\sum_{l=1}^{k-1} \alpha_{t+l}-\sqrt{\epsilon})^2 \geq 4\epsilon >\epsilon,
\end{equation*} 
which contradicts the convergence assumption. Thus, ADAM diverges for this unconstrained stochastic optimization problem. This completes the proof of Theorem~\ref{thm:DivOP}.
\subsection{Proof of Theorem~\ref{thm:DivFixb}}
Consider function $\pi(\delta) = (1+\delta)/(1+\delta^4)$. We notice that $\pi(1)=1$, $\pi(\delta)\leq 1$ for $\delta\geq 1$, and $\pi(+\infty)=0$. Since $\pi$ has only a finite number of stationary points, there exists a $\bar \delta$ such that $\pi$ is decreasing on $[\bar\delta,\infty)$. Thus for any $b$, there exists an $N^*_b$ such that for any $N\geq N^*_b, N>b$ there exists a $\delta_{N,b}>\max(\delta^*,\bar\delta)>\delta^*$ with 
\begin{equation*}
    \frac{b}{N} = \pi(\delta_{N,b}).
\end{equation*}
Let us consider the following mini-batch problem with sample size $N>N^*_b$.
\begin{eqnarray}
    f_n(w) &=& \frac{w^2}{2\delta_{N,b}} - w\;\;\;\;\text{for}\;n=1,\ldots, N-1,\nonumber\\
    f_N(w) &=& \frac{w^2}{2\delta_{N,b}} + (b\delta_{N,b}^4 + b-1)w.\nonumber
\end{eqnarray}
Apparently, the selection of mini-batch $\BB_t$ satisfies
$$\PP\{N\in\BB_t\} = \frac{\binom{N-1}{b-1}}{ \binom{N}{b} } = \frac{b}{N} = \frac{1+\delta_{N,b}}{1+\delta_{N,b}^4}:=p.$$
If $M\in\BB_t$, we have
\begin{eqnarray}
F^{\BB_t}(w) = \frac{1}{b}\left( f_{N}(w) + (b-1)f_{1}(w)\right) = \frac{w^2}{2\delta_{N,b}} +\delta_{N,b}^b w. \nonumber
\end{eqnarray}
Otherwise, it is clear that
$$F^{\BB_t}(w) = f_1(w) = \frac{w^2}{2\delta_{N,b}} - w.$$
To summarize, the mini-batch loss reads
\begin{eqnarray}
F^{\BB_t}(w) = \begin{cases}\frac{w^2}{2\delta_{N,b}} + \delta_{N,b}^4 w & \text{with probability } \;p\\
\frac{w^2}{2\delta_{N,b}} - w & \text{with probability }\; 1-p
\end{cases}\nonumber
\end{eqnarray}
which is an $\mathrm{OP}(\delta_{N,b})$, since $\delta_{N,b}>\delta^*$. By Theorem~\ref{thm:DivOP}, ADAM diverges on this problem.

\subsection{Proof of Theorem~\ref{thm:DivLargeb}}
Similarly to the proof of Theorem~\ref{thm:DivFixb}, there exists $N^*$ such that for each $N>N^*$, there exists a $\delta_N>\delta^*$ such that 
\begin{equation*}
    \frac{1}{N} = \frac{1+\delta_{N}}{1+\delta_{N}^{4}}.
\end{equation*}
We let 
\begin{eqnarray}
f_n(w) &=& \frac{w^2}{2\delta_N} + \delta_N^4 w\;\;\text{for } n=1,\ldots, N-1\nonumber\\
f_N(w) &=& \frac{w^2}{2\delta_N} -((N-1) + (N-2)\delta_N^4) w.\nonumber
\end{eqnarray}
The selection of mini-batch $\BB_t$ satisfies
\begin{eqnarray}
\PP\{N\not \in \BB_t\} = \frac{1}{N}. \nonumber
\end{eqnarray}
If $N\not \in \BB_t$, we have
\begin{equation*}
    F^{\BB_t}(w) = f_1(w) = \frac{w^2}{2\delta_{N}} + \delta_N^4 w,
\end{equation*}
and otherwise
\begin{equation*}
    F^{\BB_t}(w) = \frac{N-2}{N-1}f_1(w)+\frac{1}{N-1}f_N(w) = \frac{w^2}{2\delta_N} - w.
\end{equation*}
This is an OP($\delta_N$), which is divergent according to Theorem~\ref{thm:DivOP}.

\subsection{Proof of Theorem~\ref{thm:MtvNonCvx}}
We first introduce the following lemma.
\begin{lemma}
\label{lem:BoundStepMotiv}
Given Assumption~\ref{assmpt:Basic} is satisfied,  there exist positive constants $Q_1$ and $Q_2$ such that for any $t$, 
\begin{equation*}
    \EE[F(w_{t+1})] - \EE[F(w_t)] \leq -\frac{\alpha_t}{4\sqrt{G^2 + \epsilon}} \EE\left[\|\nabla F(w_t)\|_2^2\right] + Q_1 \alpha_t\lambda_t + Q_2 \alpha_t \sum_{k=1}^{t-1}\beta_1^{t-k}\lambda_k +Q_3 \alpha_t^2.
\end{equation*}
\end{lemma}
\textbf{Proof:}
Let us start from the application of $L$-smoothness of gradient of $F(w)$ as follows.
\begin{eqnarray}
\label{main}
F(w_{t+1}) &\leq& F(w_t) + \nabla F(w_t) ^{\top} (w_{t+1} - w_t) + \frac{L}{2}\left\|w_{t+1} - w_t\right\|_2^2  \nonumber\\
&=& F(w_t) -\frac{\alpha_t}{1-\beta_1^{t}} \nabla F(w_t) ^{\top} V_t^{-1/2} m_t + \frac{\alpha_t^2L}{2(1-\beta_1^t)^2} \left\|V_t^{-1/2} m_t \right\|_2^2 \nonumber\\
&=& F(w_t) - \frac{\alpha_t(1-\beta_1)}{1-\beta_1^t} \nabla F(w_t)^{\top} V_t^{-1/2} \sum_{k=1}^t \beta_1^{t-k} g_k + \frac{\alpha_t^2 L}{2(1-\beta_1^t)^2}\left\| V_t^{-1/2} m_t\right\|_2^2 \nonumber\\
&=& F(w_t) -  \alpha_t \nabla F(w_t)^{\top} V_t^{-1/2} g_t- \frac{\alpha_t (1-\beta_1)}{1-\beta_1^t}  \sum_{k=1}^{t-1} \beta_1^{t-k}\nabla F(w_t)^\top V_t^{-1/2}(g_k-g_t) \nonumber\\
&& +  \frac{\alpha_t^2 L}{2(1-\beta_1^t)^2} \left\| V_t^{-1/2} m_t\right\|_2^2\nonumber\\
&=& F(w_t) - \alpha_t\underbrace{ \nabla F(w_t)^{\top} V_t^{-1/2}\GG(w_t;\xi_t)}_{T_1}\nonumber\\
&& -\frac{\alpha_t(1-\beta_1)}{1-\beta_1^t}\underbrace{ \sum_{k=1}^{t-1}\beta_1^{t-k} \nabla F(w_t)^{\top} V_t^{-1/2}\left(\GG(w_k;\xi_k) - \GG(w_k;\xi_t)\right)}_{T_2}\nonumber\\
&&- \frac{\alpha_t(1-\beta_1)}{1-\beta_1^t} \underbrace{\sum_{k=1}^{t-1}\beta_1^{t-k} \nabla F(w_t)^{\top} V_t^{-1/2}\left(\GG(w_k;\xi_t) - \GG(w_t;\xi_t)\right) }_{T_3}\nonumber\\
&&+ \frac{\alpha_t^2 L}{2(1-\beta_1^t)^2}\underbrace{\left\|V_t^{-1/2}m_t\right\|_2^2}_{T_4}.\nonumber
\end{eqnarray}

\textbf{\emph{Bounding $T_1$:}} We start from
\begin{eqnarray}
\EE[T_1] &=&  \EE\left[\left\| V_t^{-1/4} \nabla F(w_t)\right\|_2^2 \right] + \EE\left[ \nabla F(w_t) ^\top V_t^{-1/2}(\GG(w_t;\xi_t) - \nabla F(w_t))\right]\nonumber\\
&\geq&  \frac{1}{2}\EE\left[\left\| V_t^{-1/4} \nabla F(w_t)\right\|_2^2 \right]-\frac{1}{2}\EE\left[ \left\| V_t^{-1/4} (\GG(w_t;\xi_t) - \nabla F(w_t))\right\|_2^2\right]\nonumber \\
&\geq& \frac{1}{2\sqrt{G^2+\epsilon}} \EE\left[\left\|\nabla F(w_t)\right\|_2^2\right] - \frac{1}{2\sqrt{\epsilon}} \EE\left[\left\|\GG(w_t;\xi_t) - \nabla F(w_t) \right\|_2^2 \right],\nonumber
\end{eqnarray}
where the first inequality applies the Cauchy-Schwartz inequality and the second inequality is due to
\begin{equation*}
\left\| V_t^{-1/4} \nabla F(w_t)\right\|_2^2 = \sum_{i=1}^d \frac{\left(\nabla_i F(w_t)\right)^2}{\sqrt{\tilde v_{t,i} + \epsilon}} \geq \frac{1}{\sqrt{G^2+\epsilon}} \sum_{i=1}^d \left(\nabla_i F(w_t)\right)^2 = \frac{1}{\sqrt{G^2 + \epsilon}} \left\| \nabla F(w_t)\right\|_2^2
\end{equation*}
and
\begin{eqnarray}
\left\| V_t^{-1/4} (\GG(w_t;\xi_t) - \nabla F(w_t))\right\|_2^2 &=& \sum_{i=1}^d \frac{\left( \nabla_i F(w_t) - \GG_i(w_t;\xi_t)\right)^2}{\sqrt{\tilde v_{t,i} + \epsilon}} \nonumber\\
&\leq& \frac{1}{\sqrt{\epsilon}}\sum_{i=1}^d\left( \nabla_i F(w_t) - \GG_i(w_t;\xi_t)\right)^2 \nonumber\\
&=& \frac{1}{\sqrt{\epsilon}}\left\| \GG(w_t;\xi_t) - \nabla F(w_t)\right\|_2^2.\nonumber
\end{eqnarray}
According to the unbiased assumption, we have
\begin{equation}
\label{BdNorm}
\EE\left[\left\|\GG(w_t;\xi_t) - \nabla F(w_t) \right\|_2^2 \right] = \sum_{i=1}^d \EE\left[(\GG_i(w_t;\xi_t) - \nabla_i F(w_t))^2\right] = \sum_{i=1}^d \mathrm{Var}(\GG_i(w_t;\xi_t)) \leq d\lambda_t.
\end{equation}
Then we can lower bound the expectation of $T_1$ as
\begin{equation}
    \label{MtvT1}
    \EE[T_1] \geq \frac{1}{2\sqrt{G^2+\epsilon}} \EE\left[\left\|\nabla F(w_t)\right\|_2^2\right] - \frac{d}{2\sqrt{\epsilon}}\lambda_t.
\end{equation}

\textbf{\emph{Bounding $T_2$:}}
Notice that for two random vectors $X$ and $Y$, and a constant $a$ we have
$$\left\|aX+\frac{1}{a}Y\right\|_2^2 = a^2\|X\|_2^2 + \frac{1}{a^2}\|Y\|_2^2 + 2 Y^\top X,$$
and thus
$$\EE\left[Y^\top X\right] \geq -\frac{a^2}{2}\EE[\|X\|_2^2] - \frac{1}{2a^2}\EE[\|Y\|_2^2].$$
If $\FF_k = \{\xi_1,\ldots,\xi_{k-1}\}$, then $w_k$ is known given $\FF_k$. We then have
\begin{eqnarray}
\EE\left[T_2\right] &=& \sum_{k=1}^{t-1} \beta_1^{t-k} \EE\left[\nabla F(w_t)^{\top} V_{t}^{-1/2}\left(\GG(w_k;\xi_k)-\GG(w_k;\xi_t) \right)\right]\nonumber\\
&=& \sum_{k=1}^{t-1} \beta_1^{t-k}\EE\left[ \EE\left[\nabla F(w_t)^{\top} V_{t}^{-1/2}\left(\GG(w_k;\xi_k)-\GG(w_k;\xi_t) \right)\left|\FF_k\right.\right]\right]\nonumber\\
&\geq& -\frac{1}{2}\sum_{k=1}^{t-1} \beta_1^{t-k} \EE\left[ a^2\EE\left[\|V_t^{-1/2}\nabla F(w_t) \|_2^2\left|\FF_k\right.\right] + \frac{1}{a^2}\EE\left[\left\|\GG(w_k;\xi_k)-\GG(w_k;\xi_t)\right\|_2^2\left|\FF_k\right.\right] \right]\nonumber\\
&\geq& -\frac{1}{2}\sum_{k=1}^{t-1} \beta_1^{t-k}\left\{\frac{a^2}{\epsilon} \EE\left[ \left\|\nabla F(w_t)\right\|_2^2\right] + \frac{1}{a^2}\EE\left[ \EE\left[ \left\|\GG(w_k;\xi_k)-\GG(w_k;\xi_t)\right\|_2^2\left|\FF_k\right.\right]\right] \right\}\nonumber\\
&=& -\frac{1}{2} \sum_{k=1}^{t-1} \beta_1^{t-k}\left\{\frac{a^2}{\epsilon} \EE\left[ \left\|\nabla F(w_t)\right\|_2^2\right] + \frac{2}{a^2}\EE\left[ \EE\left[ \left\|\nabla \GG(w_k;\xi_k) - \nabla F(w_k)\right\|_2^2\left|\FF_k\right.\right]\right] \right\}\nonumber\\
&\geq& -\frac{a^2}{2\epsilon} \frac{1}{1-\beta_1} \EE\left[ \left\|\nabla F(w_t)\right\|_2^2\right] - \frac{1}{a^2} \sum_{k=1}^{t-1} \beta_1^{t-k} \EE\left[ \left\|\nabla \GG(w_k;\xi_k) - \nabla F(w_k)\right\|_2^2\right]\nonumber\\
&\geq& -\frac{a^2}{2\epsilon} \frac{1}{1-\beta_1} \EE\left[ \left\|\nabla F(w_t)\right\|_2^2\right] - \frac{d}{a^2} \sum_{k=1}^{t-1} \beta_1^{t-k}\lambda_k\nonumber
\end{eqnarray}
for any positive constant $a$, where the third equality holds because $\GG(w_k;\xi_k)$ and $\GG(w_k;\xi_t)$ are i.i.d. given $\FF_k$, and thus
\begin{eqnarray}
    \EE\left[\left\|\GG(w_k;\xi_k)-\GG(w_k;\xi_t)\right\|_2^2\left|\FF_k\right.\right] &=& \EE\left[\left\|\GG(w_k;\xi_k)-\nabla F(w_k)\right\|_2^2\left|\FF_k\right.\right] +  \EE\left[\left\|\GG(w_k;\xi_t)-\nabla F(w_k)\right\|_2^2\left|\FF_k\right.\right]\nonumber\\
    && - 2  \EE\left[\left(\GG(w_k;\xi_k)-\nabla F(w_k)\right)\left|\FF_k\right.\right]^{\top} \EE\left[\left(\GG(w_k;\xi_t)-\nabla F(w_k)\right)\left|\FF_k\right.\right]\nonumber\\
    &=& 2 \EE\left[\left\|\GG(w_k;\xi_k)-\nabla F(w_k)\right\|_2^2\left|\FF_k\right.\right].\nonumber
\end{eqnarray}
The last inequality applies (\ref{BdNorm}).

 If $a = \sqrt{\epsilon(1-\beta_1)/2\sqrt{G^2+\epsilon}}$, then we have
\begin{eqnarray}
\label{MtvT2}
\EE[T_2] \geq -\frac{1}{4\sqrt{G^2+\epsilon}}\EE\left[\left\|\nabla F(w_t)\right\|_2^2\right] - \frac{2d\sqrt{G^2+\epsilon}}{\epsilon (1-\beta_1)}\sum_{k=1}^{t-1} \beta_1^{t-k}\lambda_k.
\end{eqnarray}

\textbf{\emph{Bounding $T_3$:}} We derive
\begin{eqnarray}
T_3 
&=& \sum_{k=1}^{t-1} \beta_1^{t-k} \nabla F(w_t) V_{t}^{-1/2} \left(\GG(w_k;\xi_t)-\GG(w_t;\xi_t) \right) \nonumber \\
&\geq& -\sum_{k=1}^{t-1} \beta_1^{t-k} \left\|\nabla F(w_t) V_{t}^{-1/2}\right\|_2 \left\|\GG(w_k;\xi_t)-\GG(w_t;\xi_t)  \right\|_2 \nonumber \\
&\geq& - \frac{LG}{\sqrt{\epsilon}}\sum_{k=1}^{t-1} \beta_1^{t-k}\|w_{t}-w_k\|\nonumber\\
&=& - \frac{LG}{\sqrt{\epsilon}}\sum_{k=1}^{t-1} \beta_1^{t-k}\left\|\sum_{j=k}^{t-1} \alpha_j V_{j}^{-1/2}\tilde m_j\right\|_2\nonumber\\
&\geq& -\frac{LG}{\sqrt{\epsilon}} \sum_{k=1}^{t-1} \beta_1^{t-k}\sum_{j=k}^{t-1} \alpha_j\left\| V_j^{-1/2} \tilde m_j\right\|_2\nonumber\\
&\geq& - \frac{LG^2}{\epsilon\sqrt{1-\beta_1}} \sum_{k=1}^{t-1}\sum_{j=k}^{t-1} \beta_1^{t-k}\alpha_j\nonumber\\
&=& -\frac{LG^2}{\epsilon\sqrt{1-\beta_1}} \sum_{j=1}^{t-1} \alpha_j \sum_{k=1}^{j} \beta_1^{t-k}\nonumber\\
&\geq& -\frac{LG^2}{\epsilon(1-\beta_1)^{3/2}}\sum_{j=1}^{t-1} \alpha_j \beta_1^{t-j}\nonumber \\
&\geq& -\frac{LG^2\bar C}{\epsilon(1-\beta_1)^{3/2}}\alpha_t,\label{MtvT3}
\end{eqnarray}
where the first inequality is the Cauchy-Schwartz inequality, the second inequality applies $L$ smoothness of $G(\cdot;\xi)$ for any $\xi$, the forth inequality holds because
\begin{eqnarray}
\left\|V_j^{-1/2} \tilde m_j\right\|_2 = \sqrt{\frac{1}{1-\beta_1^j} \sum_{i=1}^d  \frac{m_{j,i}^2}{v_{j,i} + \epsilon}}\leq \frac{G}{\sqrt{\epsilon(1-\beta_1)}}\nonumber
\end{eqnarray}
and the last inequality comes from Lemma~\ref{lem:sumbetaalpha}.

\textbf{\emph{Bounding $T_4$:}}
It is easy to show that
\begin{eqnarray}
\label{MtvT4}
T_4 = \sum_{i=1}^d \frac{m_{t,i}^2}{v_{t,i} + \epsilon} \leq \frac{G^2 }{\epsilon }.
\end{eqnarray}

According to the bounds in (\ref{MtvT1}), (\ref{MtvT2}), (\ref{MtvT3}) and (\ref{MtvT4}),we get
\begin{eqnarray}
\EE[F(w_{t+1})] - \EE[F(w_t)] &\leq& -\alpha_t  \left\{\frac{1}{2\sqrt{G^2+\epsilon}} \EE\left[\left\|\nabla F(w_t)\right\|_2^2\right] - \frac{d}{2\sqrt{\epsilon}}\lambda_t \right\} \nonumber\\
&&-\frac{\alpha_t(1-\beta_1)}{1-\beta_1^t}\left\{ -\frac{1}{4\sqrt{G^2+\epsilon}}\EE\left[\left\|\nabla F(w_t)\right\|_2^2\right] - \frac{2d\sqrt{G^2+\epsilon}}{\epsilon (1-\beta_1)}\sum_{k=1}^{t-1} \beta_1^{t-k}\lambda_k\right\} \nonumber\\
&& + \frac{LG^2\bar C}{\epsilon\sqrt{1-\beta_1}(1-\beta_1^t)}\alpha_t^2 + \frac{LG^2}{2\epsilon(1-\beta_1^t)^2} \alpha_t^2\nonumber\\
&\leq& -\alpha_t \frac{1}{4\sqrt{G^2+\epsilon}}\EE\left[\left\|\nabla F(w_t)\right\|^2_2\right] + \frac{d}{2\sqrt{\epsilon}} \alpha_t\lambda_t + \frac{2d\sqrt{G^2+\epsilon}}{\epsilon(1-\beta_1)}\alpha_t\sum_{k=1}^{t-1}\beta_1^{t-k}\lambda_k\nonumber\\
&&+\left\{\frac{LG^2\bar C}{\epsilon(1-\beta_1)^{3/2}} + \frac{LG^2}{2\epsilon(1-\beta_1)^2} \right\}\alpha_t^2.\nonumber
\end{eqnarray}
Letting
\begin{eqnarray}
Q_1&=& \frac{d}{2\sqrt{\epsilon}}\nonumber\\
Q_2&=& \frac{2d\sqrt{G^2+\epsilon}}{\epsilon(1-\beta_1)}\nonumber\\
Q_3 &=& \frac{LG^2\bar C}{\epsilon(1-\beta_1)^{3/2}} + \frac{LG^2}{2\epsilon(1-\beta_1)^2} \nonumber
\end{eqnarray}
completes the proof.

\textbf{Proof of Theorem~\ref{thm:MtvNonCvx}:} According to Lemma~\ref{lem:BoundStepMotiv}, we have
\begin{eqnarray}
F_{\mathrm{inf}} - F(w_1) &\leq& \EE[F(w_{T+1})] - F(w_1) \nonumber\\
&=& \sum_{t=1}^T \EE[F(w_{t+1})] - \EE[F(w_t)]\nonumber\\
&\leq& -\frac{1}{4\sqrt{G^2 + \epsilon}} \sum_{i=1}^T \alpha_t \EE\left[\|\nabla F(w_t)\|_2^2\right] + Q_1\sum_{i=1}^T \alpha_t\lambda_t + Q_2 \sum_{i=1}^T \alpha_t \sum_{k=1}^{t-1}\beta_1^{t-k}\lambda_k\nonumber\\
&& +Q_3 \sum_{i=1}^{T}\alpha_t^2.\nonumber
\end{eqnarray}
Then we obtain
\begin{eqnarray}
\frac{1}{4\sqrt{G^2+\epsilon}}\sum_{t=1}^T \alpha_t \EE\left[\|\nabla F(w_t)\|_2^2\right] &\leq& F(w_1) - F_{\mathrm{inf}} +Q_1 \sum_{t=1}^T \alpha_t\lambda_t \nonumber\\
&&+Q_2 \sum_{t=1}^T \alpha_t \sum_{k=1}^{t-1}\beta_1^{t-k}\lambda_k + Q_3\sum_{t=1}^T \alpha_t^2\nonumber\\
&\leq&F(w_1) - F_{\mathrm{inf}} +Q_1 \sum_{t=1}^T \alpha_t\lambda_t \nonumber\\
&&+Q_2 \sum_{k=1}^T \lambda_k \sum_{t=k}^{T}\beta_1^{t-k}\alpha_t +  Q_3\sum_{t=1}^T \alpha_t^2 \nonumber\\
&\leq &F(w_1) - F_{\mathrm{inf}} +Q_1 \sum_{t=1}^T \alpha_t\lambda_t \nonumber\\
&&+\frac{Q_2}{1-\beta_1} \sum_{k=1}^T \lambda_k \alpha_k +  Q_3\sum_{t=1}^T \alpha_t^2 .\nonumber
\end{eqnarray}
Noticing that the left-hand side can be bounded as
\begin{equation}
    \sum_{t=1}^T \alpha_t \EE\left[\left\|\nabla F(w_t)\right\|_2^2\right] \geq \sum_{t=1}^T \alpha_t \min_{1\leq t\leq T} \EE\left[\left\|\nabla F(w_t)\right\|_2^2\right],\nonumber
\end{equation}
we obtain
\begin{eqnarray}
\min_{1\leq t \leq T} \EE\left[\left\|\nabla F(w_t)\right\|_2^2\right] &\leq& \frac{4\sqrt{G^2+\epsilon}}{\sum_{t=1}^T \alpha_t} + 4\sqrt{G^2+\epsilon}\left(Q_1 + \frac{Q_2}{1-\beta_2}\right)\frac{\sum_{t=1}^T \lambda_t\alpha_t}{\sum_{t=1}^T\alpha_t}\nonumber\\
&&+ 4\sqrt{G^2+\epsilon} Q_3 \frac{\sum_{t=1}^T \alpha_t^2}{\sum_{t=1}^T \alpha_t}\nonumber.
\end{eqnarray}

\subsection{Proof of Theorem~\ref{thm:ResetBetter}}
 At the end of $t=1$ iteration, we obtain $w_{m+1}^{(1)} = w_1^{(2)}=\tilde w_2$ and the ADAM states $m_{m+1}^{(1)}$ and $v_{m+1}^{(1)}$. Then while $t=2$, the algorithm makes the first update as follows.
\begin{equation*}
    \begin{split}
    &\textbf{Option A:}\\
        m^{(2)}_1 &= (1-\beta_1) g^{(2)}_1\\
        \widetilde m^{(2)}_1 &= g^{(2)}_1\\
        v^{(2)}_1 &= (1-\beta_2) \left(g^{(2)}_1\right)^2\\ \widetilde v^{(2)}_1 &= \left(g^{(2)}_1\right)^2\\
        w^{(2)}_2 &= w^{(2)}_1 - \alpha_2 \frac{\widetilde m^{(2)}_1}{\sqrt{\widetilde v^{(2)}_1 + \epsilon}}
    \end{split}
    \hspace{1cm}
    \begin{split}
    &\textbf{Option B:}\\
        \hat m^{(2)}_1 &=\beta_1 m^{(1)}_{m+1} + (1-\beta_1) g^{(2)}_1\\
        \hat{ \widetilde m}^{(2)}_1 &= \hat m^{(2)}_1 / (1-\beta_1^{m+1}) \\
        \hat v^{(2)}_1&= \beta_2 v^{(1)}_{m+1} + (1-\beta_2)\left( g^{(2)}_1\right)^2\\
        \hat{ \widetilde v}_1^{(2)} &= \hat v_1^{(2)}/(1-\beta_2^{m+1}) \\
        \hat w_2^{(2)} &=  w^{(2)}_1 - \alpha_2 \frac{ \hat{\widetilde m}^{(2)}_1}{\sqrt{\hat{\widetilde v}_1^{(2)}+\epsilon}}
    \end{split}
\end{equation*}

Applying L-smoothness of gradients of $F$ and strong convexity of $F$, we have
\begin{eqnarray}
F\left(\hat w^{(2)}_2\right) &\geq& F\left(w^{(2)}_1\right) + F'\left(w^{(2)}_1\right)\left(\hat w^{(2)}_2 - w^{(2)}_1\right) + \frac{c}{2}\left(\hat w^{(2)}_2 - w^{(2)}_1\right)^2\nonumber\\
F\left( w^{(2)}_2\right) &\leq& F\left(w^{(2)}_1\right) + F'\left(w^{(2)}_1\right)\left( w^{(2)}_2 - w^{(2)}_1\right) + \frac{L}{2}\left( w^{(2)}_2 - w^{(2)}_1\right)^2.\nonumber
\end{eqnarray}
By definition, we have
\begin{eqnarray}
F\left(\hat w^{(2)}_2\right) - F\left( w^{(2)}_2\right) &\geq& F'\left(w^{(2)}_1\right)\left(\hat w^{(2)}_2 - w^{(2)}_2\right) + \frac{c}{2}\left(\hat w^{(2)}_2 - w^{(2)}_1\right)^2 - \frac{L}{2}\left( w^{(2)}_2 - w^{(2)}_1\right)^2\nonumber\\
&=& F'\left(w^{(2)}_1\right) \left( \alpha_2\frac{\widetilde m^{(2)}_1}{\sqrt{\widetilde v^{(2)}_1 + \epsilon}} - \alpha_2\frac{\hat{\widetilde m}^{(2)}_1}{\sqrt{\hat{\widetilde v}^{(2)}_1 + \epsilon}}\right) + \frac{c\alpha_2^2}{2} \frac{\hat{\widetilde m}^{(2)}_1}{\hat{\widetilde v}^{(2)}_1 + \epsilon} - \frac{L\alpha_2^2}{2} \frac{{\widetilde m}^{(2)}_1}{{\widetilde v}^{(2)}_1 + \epsilon}\nonumber\\
&=& \alpha_2  F'\left(w^{(2)}_1\right) \left( \frac{ F'\left(w^{(2)}_1\right)}{\sqrt{Q_3}} -\gamma \frac{ (1-\beta_1)F'\left(w^{(2)}_1\right) + \beta_1 m^{(1)}_{m+1} }{\sqrt{Q_4}}\right) \nonumber\\
&& + \frac{c\alpha_2^2\gamma^2}{2}\frac{ \left( (1-\beta_1)F'\left(w^{(2)}_1\right) + \beta_1 m^{(1)}_2 \right)^2}{Q_4} - \frac{L\alpha_2^2}{2}  \frac{ \left(F'\left(w^{(2)}_1\right)\right)^2}{Q_3} \nonumber
\end{eqnarray}
where
\begin{eqnarray}
Q_3 &=& \widetilde v^{(2)}_1 + \epsilon \nonumber\\
Q_4 &=& \hat{\widetilde v}^{(2)}_1 + \epsilon \nonumber\\
\gamma &=& \frac{1}{1-\beta_1^{m+1}}.\nonumber
\end{eqnarray}
Thus we have
\begin{eqnarray}
F\left(\hat w^{(2)}_2\right) - F\left( w^{(2)}_2\right)& \geq &  \left(F'\left(w^{(2)}_1\right)\right)^2 q\left( \frac{m^{(1)}_{m+1}}{F'\left(w^{(2)}_{1}\right)}\right)\nonumber
\end{eqnarray}
where $q(x) = Q_5 x^2 + Q_6 x + Q_7$ is a function with parameters
\begin{eqnarray}
Q_5 &=& \frac{c\alpha_2^2\gamma^2 \beta_1^2}{2Q_4} \nonumber\\
Q_6 &=& \frac{c\alpha_2^2\gamma^2\beta_1(1-\beta_1)}{Q_4}  - \frac{\alpha_2 \gamma \beta_1}{\sqrt{Q_2}}\nonumber\\
Q_7 &=& \frac{\alpha_2}{\sqrt{Q_3}} - \frac{\alpha_2\gamma(1-\beta_1)}{\sqrt{Q_4}} + \frac{c\alpha_2^2\gamma^2(1-\beta_1)^2}{2Q_4} - \frac{L\alpha_2^2}{2Q_3}.\nonumber
\end{eqnarray}
Apparently, from
\begin{eqnarray}
\widetilde v_1^{(2)} &=& \left(g_1^{(2)}\right)^2\leq G^2\nonumber\\
\hat{\widetilde v}_1^{(2)} &=& \frac{1-\beta_1}{1-\beta_2^{m+1}} \left(\sum_{k=1}^m \beta_2^{m + 1 - k} \left(g_k^{(1)}\right)^2 + \left(g_1^{(2)}\right)^2\right) \leq \frac{1-\beta_1}{1-\beta_2^{m+1}} \left(\sum_{k=1}^m \beta_2^{m + 1 - k}  + 1\right)G^2 = G^2\nonumber,
\end{eqnarray}
we have
\begin{eqnarray}
\epsilon \leq &Q_3& \leq \epsilon + G^2\nonumber\\
\epsilon \leq &Q_4& \leq \epsilon + G^2.\nonumber
\end{eqnarray}

Noticing that 
\begin{eqnarray}
\Delta &=& Q_6^2 - 4Q_5Q_7\nonumber\\
&=& \frac{\alpha_2^2\gamma^2\beta_1^2}{Q_4}\left(1 - \frac{2c\alpha_2}{\sqrt{Q_3}} + \frac{cL\alpha_2^2}{Q_3}\right)\nonumber\\
&>& \frac{\alpha_2^2\gamma^2 \beta_1^2}{Q_4}\left(1 - \frac{2c\alpha_2}{\sqrt{Q_3}} + \frac{c^2\alpha_2^2}{Q_3}\right)\nonumber\\
&=&\frac{\alpha_2^2\gamma^2 \beta_1^2}{Q_4}\left(1-\frac{c\alpha_2}{\sqrt{Q_3}}\right)^2\geq 0, \nonumber
\end{eqnarray}
where the first inequality uses the property of the strong convexity parameter and the L-smoothness gradient parameter $c<L$, we have that there exists 
\begin{eqnarray}
x_1 &=& \frac{-Q_6 + \sqrt{\Delta}}{2Q_5}\nonumber\\
x_2 &=& \frac{-Q_6 - \sqrt{\Delta}}{2Q_5} \nonumber
\end{eqnarray}
such that $q(x_1)=q(x_2)=0$. We claim that $|x_1|\leq 1$ and $|x_2|\leq 1$, which is implied by
\begin{equation}
\label{BdSqrtDelta}
\sqrt{\Delta} \leq \min\{ 2Q_5 + Q_6, 2Q_5 - Q_6\}.
\end{equation}
We notice that
\begin{eqnarray}
2Q_5 + Q_6 &=& \frac{\alpha_2 \gamma \beta_1}{\sqrt{Q_4}}\left( \frac{c
\alpha_2 \gamma}{\sqrt{Q_4}} -1\right)\nonumber\\
2Q_5 - Q_6 &=&  \frac{\alpha_2 \gamma \beta_1}{\sqrt{Q_4}}
\left(\frac{c\alpha_2\gamma(2\beta_1-1)}{\sqrt{Q_4}} +1\right)\nonumber
\end{eqnarray}
and
\begin{eqnarray}
\Delta &\leq& \frac{\alpha_2^2 \gamma^2 \beta_1^2}{Q_4}\left(1 - \frac{2L\alpha_2}{\sqrt{Q_3}} + \frac{L^2 \alpha_2^2}{Q_3}\right)\nonumber\\
&=& \frac{\alpha_2^2 \gamma^2 \beta_1^2}{Q_4}\left(1-\frac{L\alpha_2}{\sqrt{Q_3}}\right)^2\nonumber
\end{eqnarray}
where the inequality holds according to Assumption~\ref{assmpt:Noreset} 
$
    L\alpha_2 \geq 2\sqrt{G^2 + \epsilon} \geq 2 \sqrt{Q_3}.
$
Thus we have
\begin{equation*}
    \sqrt{\Delta} \leq \frac{\alpha_2\gamma\beta_1}{\sqrt{Q_4}} \left(\frac{L\alpha_2}{\sqrt{Q_3}} - 1\right)
    \leq \frac{\alpha_2\gamma\beta_1}{\sqrt{Q_4}} \left(\frac{c\alpha_2\gamma(2\beta_1-1)}{\sqrt{Q_4}} - 1\right)
    \leq \min\{2Q_5+Q_6, 2Q_5 - Q_6\},
\end{equation*}
where the second inequality holds according to Assumption~\ref{assmpt:Noreset}, $
\frac{L}{c} \leq \frac{2\beta_1-1}{1-\beta_1^{m+1}} \sqrt{\frac{\epsilon}{G+\epsilon}} \leq \frac{2\beta_1-1}{1-\beta_1^{m+1}} \sqrt{\frac{Q_3}{Q_4}}$.

Therefore we obtain (\ref{BdSqrtDelta}), which implies that $q(x)\geq 0$ where $|x|\geq1$. As we assume
\begin{equation*}
   \left|m^{(1)}_{m+1}\right| \geq \left|F'\left(w^{(2)}_{1}\right)\right|,
\end{equation*}
we have
\begin{equation*}
    F\left(\hat w^{(2)}_2\right) - F\left( w^{(2)}_2\right)\geq 0,
\end{equation*}
which finishes the proof.

\subsection{Proof of Lemma~\ref{lem:BoundGradient}, Theorem~\ref{thm:VRStrCvx} and Theorem~\ref{thm:VRNonCvx}}

\textbf{Proof of Lemma~\ref{lem:BoundGradient}:} For all $t$, by the algorithm we have $\|\widetilde w_{t}\|_2 \leq M$. For every $1\leq k\leq m+1$ and $t$, we have
\begin{eqnarray}
    \left\|\left( V_k^{(t)}\right)^{-1/2} \widetilde m_k^{(t)}\right\|_2^2 &=& \sum_{i=1}^d \frac{\left(\widetilde m_{k,i}^{(t)}\right)^2}{\widetilde v_{k,i}^{(t)} + \epsilon} 
    = \sum_{i=1}^d \frac{\frac{1}{1-\beta_1^k} \left(\sum_{j=1}^k \beta_1^{k-j} g_{j,i}^{(t)}\right)^2}{\frac{1}{1-\beta_2^k} \sum_{l=1}^k \beta_2^{k-l} \left(g_{l,i}^{(t)}\right)^2+\epsilon}\nonumber\\
    &\leq& \frac{1-\beta_2^k}{1-\beta_1^k}\sum_{i=1}^d \frac{ \left(\sum_{j=1}^k \beta_1^{k-j} g_{j,i}^{(t)}\right)^2}{ \sum_{l=1}^k \beta_2^{k-l} \left(g_{l,i}^{(t)}\right)^2}\nonumber\\
    &\leq& k\frac{1-\beta_2^k}{1-\beta_1^k}\sum_{i=1}^d \frac{ \sum_{j=1}^k \beta_1^{2(k-j)} \left(g_{j,i}^{(t)}\right)^2}{ \sum_{l=1}^k \beta_2^{k-l} \left(g_{l,i}^{(t)}\right)^2}\nonumber\\
    &=& k\frac{1-\beta_2^k}{1-\beta_1^k}\sum_{i=1}^d \sum_{j=1}^k \left(\frac{\beta_1^2}{\beta_2}\right)^{k-j} \frac{  \beta_2^{k-j} \left(g_{j,i}^{(t)}\right)^2}{ \sum_{l=1}^k \beta_2^{k-l} \left(g_{l,i}^{(t)}\right)^2}\nonumber\\
    &\leq & k\frac{1-\beta_2^k}{1-\beta_1^k}\sum_{i=1}^d \sum_{j=1}^k \left(\frac{\beta_1^2}{\beta_2}\right)^{k-j}= kd\frac{1-\beta_2^k}{1-\beta_1^k} \sum_{j=0}^{k-1} \left(\frac{\beta_1^2}{\beta_2}\right)^{j}\nonumber\\
    &\leq& \frac{(m+1)d}{1-\beta_1} \sum_{j=0}^{m} \left(\frac{\beta_1^2}{\beta_2}\right)^{j}:=\Delta.\nonumber
\end{eqnarray}
Clearly $\Delta = \frac{(m+1)d}{1-\beta_1}\frac{1-(\beta_1^2/\beta_2)^{m+1}}{1-\beta_1^2/\beta_2} $ for $\beta_1^2\neq \beta_2$ and $\Delta = \frac{(m+1)^2d}{1-\beta_1}$ for $\beta_1^2 = \beta_2$.

Then, by using the triangular inequality, we have 
\begin{eqnarray}
   \left\|w_{k}^{(t)}\right\|_2 &=& \left\|\widetilde w_t +\alpha_t \sum_{j=1}^{k-1} \left(V_{j}^{(t)}\right)^{-1/2} \widetilde m_j^{(t)} \right\|_2\nonumber\\
   &\leq& \left\|\widetilde w_t\right\|_2 +\alpha_t\sum_{j=1}^{k-1}\left\|  \left(V_{j}^{(t)}\right)^{-1/2} \widetilde m_j^{(t)} \right\|_2\nonumber\\
   &\leq& M + \alpha (k-1) \Delta \leq M+m\alpha\Delta :=D\nonumber.
\end{eqnarray}
Noticing that $D$ is only dependent on $d, L, M, \alpha, \beta_1$, and $\beta_2$, we have that $\left\|w_k^{(t)}\right\|\leq D$ holds for all $t=1,\ldots, T$ and $k=1,\ldots,m+1$. 

Since being Lipschitz implies continuity, $\left\|\nabla f_n(w)\right\|_2$ is uniformly upper bounded by some constant $G_n$ within the compact set $\left\{w:\|w\|_2\leq D\right\}$. Letting $G = \max_{1\leq n\leq N} G_n$ finishes the proof.
\color{black}
\begin{lemma}
\label{lem:BdMV}
For any $1\leq k\leq m$ and $1\leq t\leq T$, given that $\left\|\nabla f_n\left(w_k^{(t)}\right)\right\|_2\leq G$ holds for all $1\leq n \leq N$, the ADAM states in Algorithm~\ref{alg:VRADAM} with option A satisfy
\begin{eqnarray}
\left\|m_k^{(t)}\right\|_2 &\leq& 3G,\nonumber\\
\left\|v_k^{(t)}\right\|_2 &\leq& 9G^2.\nonumber
\end{eqnarray}
\end{lemma}

\textbf{Proof:} By definition, we have
\begin{eqnarray}
m_k^{(t)} &=& (1-\beta_1) \sum_{j=1}^{k}\beta_1^{k-j} g_j^{(t)}\nonumber\\
v_k ^{(t)} &=& (1-\beta_2) \sum_{j=1}^{k} \beta_2^{k-j} g_j^{(t)}\odot g_j^{(t)}.\nonumber
\end{eqnarray}
Applying the Cauchy-Schwartz inequality, we obtain
\begin{eqnarray}
\left\|m_k^{(t)}\right\|_2 &\leq& (1-\beta_1) \sum_{j=1}^k \beta_1^{k-j} \left\|g_j^{(t)}\right\|_2 \nonumber\\
&\leq& (1-\beta_1) \sum_{j=1}^k \beta_1^{k-j} \left(\left\|\nabla F^{\BB_j^{(t)}}\left(w_j^{(t)}\right)\right\|_2 + \left\|\nabla F^{\BB_j^{(t)}}\left(\widetilde w_t\right)\right\|_2 + \left\|\nabla F\left(\widetilde w_t\right)\right\|_2\right)\nonumber\\
&\leq& (1-\beta_1) \sum_{j=1}^k\beta_1^{k-j} 3G \leq 3G\nonumber
\end{eqnarray}
and
\begin{eqnarray}
\left\|v_k^{(t)}\right\|_2 &\leq& (1-\beta_2)\sum_{j=1}^k \beta_2^{k-j} \left\|g_j^{(t)}\odot g_j^{(t)}\right\|_2 \nonumber\\
&=&  (1-\beta_2)\sum_{j=1}^k \beta_2^{k-j} \left\|g_j^{(t)}\right\|_2^2 \nonumber\\
&\leq&(1-\beta_2) \sum_{j=1}^k \beta_2^{k-j} \left(\left\|\nabla F^{\BB_j^{(t)}}\left(w_j^{(t)}\right)\right\|_2 + \left\|\nabla F^{\BB_j^{(t)}}\left(\widetilde w_t\right)\right\|_2 + \left\|\nabla F\left(\widetilde w_t\right)\right\|_2\right)^2\nonumber\\
&\leq&(1-\beta_2)\sum_{j=1}^k \beta_2^{k-j} 9G^2 \leq 9G^2.\nonumber
\end{eqnarray}

\begin{lemma}
\label{lem:ProjectionDecrease}
Given (1) and (3) in Assumption~\ref{assmpt:LSmoothGrad}, for any $t$ we have
\begin{eqnarray}
F\left(\widetilde w_{t+1}\right) \leq F\left(w_{m+1}^{(t)} \right).\nonumber
\end{eqnarray}
\end{lemma}
\textbf{Proof:} Let $t_1 = \left\|\widetilde w_{t+1} - w^*\right\|_2$ and $t_2 = \left\|w^{(t)}_{m+1} - w^*\right\|_2$. If $\|w_{m+1}^{(t)}\|_2\leq M$, $\widetilde w_{t+1} = w_{m+1}^{(t)}$ and the proof is finished. Otherwise, we have $\widetilde w_{t+1} = M' w_{m+1}^{(t)}/\left\|w_{m+1}^{(t)}\right\|_2^2$, where $M'=\min\left\{M, \frac{c^2}{c^2+L^2} \left\| w_{m+1}^{(t)}\right\|_2\right\}.$ We start by bounding $t_1/t_2$. Notice that
\begin{eqnarray}
    \frac{t_1^2}{t_2^2} &=& \frac{ \left\| \widetilde w_{t+1} - w^*\right\|_2^2}{\left\| w^{(t)}_{m+1} - w^*\right\|_2^2}\nonumber\\
    &=& \frac{\left\|\widetilde w_{t+1} \right\|_2^2 + \left\|w^* \right\|_2^2 - 2  \widetilde w_{t+1}^\top w^*}{\left\| w_{m+1}^{(t)} \right\|_2^2 + \left\|w^* \right\|_2^2 - 2  \left(w_{m+1}^{(t)}\right)^\top w^*}\nonumber\\
    &=& \frac{M'^2+ \left\|w^* \right\|_2^2 - 2  \frac{M'}{\left\|w_{m+1}^{(t)}\right\|_2}\left(w_{m+1}^{(t)}\right)^\top w^*}{\left\| w_{m+1}^{(t)} \right\|_2^2 + \left\|w ^*\right\|_2^2 - 2  \left(w_{m+1}^{(t)}\right)^\top w^*}\nonumber\\
    &=& \frac{M'^2+ \left\|w^* \right\|_2^2 -M'\left\|w_{m+1}^{(t)}\right\|_2 -\frac{M'\left\|w^*\right\|_2^2}{\left\|w_{m+1}^{(t)}\right\|_2} +  \frac{M'}{\left\|w_{m+1}^{(t)}\right\|_2}\left[ \left\|w_{m+1}^{(t)}\right\|_2^2 + \left\|w^*\right\|_2^2-2\left(w_{m+1}^{(t)}\right)^\top w^*\right]}{\left\| w_{m+1}^{(t)} \right\|_2^2 + \left\|w^* \right\|_2^2 - 2  \left(w_{m+1}^{(t)}\right)^\top w^*} \nonumber\\
    &=& \frac{\left( 1-\frac{M'}{\left\| w_{m+1}^{(t)}\right\|_2}\right)\left(\left\| w^*\right\|_2^2 -M'\left\|w_{m+1}^{(t)}\right\|_2 \right) +  \frac{M'}{\left\|w_{m+1}^{(t)}\right\|_2}\left[ \left\|w_{m+1}^{(t)}\right\|_2^2 + \left\|w^*\right\|_2^2-2\left(w_{m+1}^{(t)}\right)^\top w^*\right]}{\left\| w_{m+1}^{(t)} \right\|_2^2 + \left\|w^* \right\|_2^2 - 2  \left(w_{m+1}^{(t)}\right)^\top w^*} \nonumber\\
    &=& \frac{M'}{\left\|w_{m+1}^{(t)}\right\|_2} + \left(1-\frac{M'}{\left\|w_{m+1}^{(t)}\right\|_2}\right) \frac{\left\|w^*\right\|_2^2 - M' \left\|w_{m+1}^{(t)}\right\|_2}{\left\|w_{m+1}^{(t)} - w^*\right\|_2^2}.\nonumber
\end{eqnarray}
If $M\geq\frac{c^2}{L^2+c^2} \left\|w_{m+1}^{(t)}\right\|_2$, we have that $M' \left\|w_{m+1}^{(t)}\right\|_2 = \frac{c^2}{L^2+c^2} \left\|w_{m+1}^{(t)}\right\|_2^2 >\frac{c^2}{L^2+c^2} M^2> \left\|w^*\right\|_2^2$, according to (3) in Assumption~\ref{assmpt:LSmoothGrad}. Otherwise, $M < \frac{c^2}{L^2+c^2} \left\|w_{m+1}^{(t)}\right\|_2^2$, and thus
$M' \left\|w_{m+1}^{(t)}\right\|_2 = M \left\|w_{m+1}^{(t)}\right\|_2 >M^2 >\left\|w^*\right\|_2^2$. Thus in either case, we have $\left\|w^*\right\|_2^2 \leq M' \left\|w_{m+1}^{(t)}\right\|_2 $ which implies that
\begin{eqnarray}
    \frac{t_1^2}{t_2^2} &\leq& \frac{M'}{\left\|w_{m+1}^{(t)}\right\|_2}\nonumber \\
    &\leq&\frac{c^2}{L^2+c^2}<1.\nonumber
\end{eqnarray}
Letting $a_1 = \frac{\widetilde w_{t+1} - w^*}{t_1}$ and $a_2 = \frac{ w_{m+1}^{(t)} - w^*}{t_2}$, we have
\begin{eqnarray}
    F\left(w_{m+1}^{(t)}\right) - F(\widetilde w_{t+1}) &=& F(w^* + t_2 a_2) - F(w^* + t_1 a_1) \nonumber\\
    &=& F(w^*+t_2 a_2) -F(w^*+t_1a_2) + F(w^*+t_1a_2) -F(w^*+t_1a_1) .\nonumber\\
\end{eqnarray}
By applying Lemma~\ref{lem:StrCvxDiff}, we obtain:
\begin{eqnarray}
     F\left(w_{m+1}^{(t)}\right) - F(\widetilde w_{t+1}) &\geq& \frac{c}{2}(t_2^2 - t_1^2) - \frac{L^2}{2c}t_1^2 \nonumber\\
     &=& \frac{c}{2} \left( t_2^2 - \frac{L^2+c^2}{c^2}t_1^2\right) \geq 0,
\end{eqnarray}
which finishes the proof.
\color{black}
\begin{lemma}
\label{lem:BoundStepVR}
Given (1) and (3) in Assumption~\ref{assmpt:LSmoothGrad}, and assume that $\left\|\nabla f_n\left(w_k^{(t)}\right)\right\|_2\leq G$ holds for all $n,k$ and $t$, there exist positive constants $Q_8$ and $Q_9$ such that Algorithm~\ref{alg:VRADAM} with option A satisfies that for any $t$, 
\begin{equation*}
F\left( w^{(t)}_{m+1} \color{black}\right) - F(\widetilde w_t) \leq - Q_8 \alpha_t m \left\|\nabla F(\widetilde w_t)\right\|_2^2 + Q_9 \alpha_t^2
\end{equation*}
holds almost surely.
\end{lemma}
\textbf{Proof:}
We start from employing $L$-smoothness of gradient of $F(w)$ as follows
\begin{eqnarray}F({w}_{m+1}^{(t)}) 
&\leq& F(\widetilde{w}_t) + \nabla F(\widetilde{w}_t) ^\top \left({w}_{m+1}^{(t)} - \widetilde{w}_t\right) + \frac{L}{2}\left\|{w}_{m+1}^{(t)} - \widetilde{w}_t\right\|_2^2 \nonumber\\
&=& F(\widetilde{w}_t) + \nabla F(\widetilde{w}_t) ^\top \sum_{k=1}^{m} \left(w^{(t)}_{k+1} - w^{(t)}_{k}\right) + \frac{L}{2}\left\|\sum_{k=1}^{m} \left(w^{(t)}_{k+1} - w^{(t)}_{k}\right)\right\|_2^2\nonumber \\
&=& F(\widetilde w_{t}) -\alpha_t \nabla F(\widetilde w_t)^\top\sum_{k=1}^{m} \left(V_k^{(t)}\right)^{-1/2} \widetilde m_k^{(t)} + \frac{\alpha_t^2L}{2}\left\|\sum_{k=1}^{m}\left(V_k^{(t)}\right)^{-1/2} \widetilde m_k^{(t)}\right\|_2^2.\nonumber
\end{eqnarray}
By definition and the resetting option, we have
\begin{eqnarray}
\widetilde m_k^{(t)} = \frac{1-\beta_1}{1-\beta_1^k} \left(V_k^{(t)}\right)^{-1/2} \sum_{j=1}^k \beta_1^{k-j} g_j^{(t)}\nonumber,
\end{eqnarray}
and thus
\begin{eqnarray}
F\left( w^{(t)}_{m+1}\right)&\leq& F(\widetilde w_t) - \alpha_t(1-\beta_1) \nabla F(\widetilde w_t)^{\top}\sum_{k=1}^{m}\frac{1}{1-\beta_1^{k}} \left(V_k^{(t)}\right)^{-1/2} \sum_{j=1}^k \beta_1^{k-j} g_j^{(t)}\nonumber\\
&&+ \frac{\alpha_t^2L}{2}\left\|\sum_{k=1}^{m}\left(V_k^{(t)}\right)^{-1/2} \widetilde m_k^{(t)}\right\|_2^2\nonumber\\ 
&=&  F(\widetilde w_t) - \alpha_t (1-\beta_1)\nabla F(\widetilde w_t)^{\top}\sum_{j=1}^{m} \left(\sum_{k=j}^{m} \frac{\beta_1^{k-j}}{1-\beta_1^k} \left(V_k^{(t)}\right)^{-1/2}\right) g_j^{(t)} 
\nonumber\\
&& +  \frac{\alpha_t^2L}{2}\left\|\sum_{k=1}^{m} \left(V_k^{(t)}\right)^{-1/2} \widetilde m_k^{(t)}\right\|_2^2\nonumber\\
&=& F(\widetilde w_t) -\alpha_t (1-\beta_1)\nonumber\\
&&\times \underbrace{\nabla F(\widetilde w_t)^{\top}\sum_{j=1}^{m} \left(\sum_{k=j}^{m} \frac{\beta_1^{k-j}}{1-\beta_1^k} \left(V_k^{(t)}\right)^{-1/2}\right) \left(\nabla F^{\BB^{(t)}_j}\left(w_j^{(t)}\right) - \nabla F^{\BB^{(t)}_j} \left(\widetilde w_t\right)\right)}_{T_1} \nonumber\\
&& - \alpha_t(1-\beta_1) \underbrace{\nabla F(\widetilde w_t)^{\top}\left(\sum_{j=1}^{m} \sum_{k=j}^{m} \frac{\beta_1^{k-j}}{1-\beta_1^k} \left(V_k^{(t)}\right)^{-1/2}\right) \nabla F(\widetilde w_t)}_{T_2}\nonumber\\
&& +  \frac{\alpha_t^2L}{2}\underbrace{\left\|\sum_{k=1}^{m} \left(V_k^{(t)}\right)^{-1/2} \widetilde m_k^{(t)}\right\|_2^2}_{T_3} .\nonumber
\end{eqnarray}
\textbf{\emph{Bounding $T_1$}}: We have
\begin{eqnarray}
\label{T10}
T_1 &\geq& -  \sum_{j=1}^{m} \left\| \sum_{k=j}^{m} \frac{\beta_1^{k-j}}{1-\beta_1^k} \left(V_k^{(t)}\right)^{-1/2}\nabla F(\widetilde w_t)\right\|_2\left\|\nabla F^{\BB^{(t)}_j}\left(w_j^{(t)}\right) - F^{\BB^{(t)}_j}(\widetilde w_t)\right\|_2.
\end{eqnarray}
The first thing to notice is that 
\begin{eqnarray}
\label{T11}
\left\| \sum_{k=j}^{m} \frac{\beta_1^{k-j}}{1-\beta_1^k} \left(V_k^{(t)}\right)^{-1/2}\nabla F(\widetilde w_t)\right\|_2&\leq& \sum_{k=j}^{m} \frac{\beta_1^{k-j}}{1-\beta_1^k}\left\| \left(V_k^{(t)}\right)^{-1/2}\nabla F(\widetilde w_t)\right\|_2\nonumber\\
&=&\sum_{k=j}^{m} \frac{\beta_1^{k-j}}{1-\beta_1^k}\sqrt{\sum_{i=1}^d \frac{\nabla_i F(\widetilde w_t)^2 }{v_{k,i}^{(t)} +\epsilon}} \nonumber\\
&\leq& \sum_{k=j}^{m} \frac{\beta_1^{k-j}}{1-\beta_1^k} \frac{\left\|\nabla F(\widetilde w_t)\right\|_2}{\sqrt{\epsilon}}\nonumber\\
&\leq& \frac{G}{\sqrt{\epsilon}}\sum_{k=j}^{m}\frac{\beta_1^{k-j}}{1-\beta_1^k} \nonumber\\
&\leq& \frac{G}{\sqrt{\epsilon}} \frac{1}{1-\beta_1}\sum_{k=j}^{m} \beta_1^{k-j}\leq \frac{G}{(1-\beta_1)^2\sqrt{\epsilon}} ,
\end{eqnarray}
where the second inequality employs the assumption that the gradients of $F$ are bounded. Secondly, according to $L$-smoothness of gradients of every loss function, we derive
\begin{eqnarray}
\label{T12}
\left\|\nabla F^{\BB^{(t)}_j}\left(w_j^{(t)}\right) - \nabla F^{\BB^{(t)}_j}\left(\widetilde w_t\right)\right\|_2 &\leq& L\left\|w_{j}^{(t)} - w_1^{(t)}\right\|_2 \nonumber\\
&=& L \left\| \sum_{l=1}^{j-1} \alpha_t \left(V_l^{(t)}\right)^{-1/2} \widetilde m_l^{(t)}\right\|_2\nonumber\\
&\leq& \alpha_t L \sum_{l=1}^{j-1} \left\| \left(V_l^{(t)}\right)^{-1/2} \widetilde m_l^{(t)}\right\|_2 \nonumber\\
&=& \alpha_t L \sum_{l=1}^{j-1} \frac{1}{1-\beta_1^l} \sqrt{\sum_{i=1}^d \frac{\left(m_{l,i}^{(t)}\right)^{2}}{v_{l,i}^{(t)}+\epsilon }}\nonumber\\
&\leq& \frac{\alpha_t L}{1-\beta_1}\sum_{l=1}^{j-1} \frac{\left\|m_l^{(t)}\right\|_2}{\sqrt{\epsilon}}\leq \frac{3GL}{(1-\beta_1)\sqrt{\epsilon}} (j-1) \alpha_t ,
\end{eqnarray}
where the first inequality applies the Cauchy-Schwartz inequality and the last one applies Lemma~\ref{lem:BdMV}. By plugging equations (\ref{T11}) and (\ref{T12}) into equation (\ref{T10}), we obtain
\begin{eqnarray}
\label{T1}
T_1 &\geq& - \sum_{j=1}^{m} \frac{3G^2L}{(1-\beta_1)^3\epsilon} (j-1) \alpha_t =- \frac{3G^2L}{2(1-\beta_1)^3\epsilon} m(m-1) \alpha_t.
\end{eqnarray}
\textbf{\emph{Bounding $T_2$}}: We have
\begin{eqnarray}
% \label{T2}
T_2 &=& \sum_{j=1}^{m}\sum_{k=j}^{m} \frac{\beta_1^{k-j}}{1-\beta_1^k} \nabla F(\widetilde w_t)^{\top} \left(V_k^{(t)}\right)^{-1/2} \nabla F(\widetilde w_t)\nonumber\\
&=&  \sum_{j=1}^{m}\sum_{k=j}^{m} \frac{\beta_1^{k-j}}{1-\beta_1^k} \sum_{i=1}^d \frac{\nabla_i F(\widetilde w_t)^2}{\sqrt{v_{k,i}^{(t)} + \epsilon}}\nonumber\\
&\geq& \sum_{j=1}^{m}\sum_{k=j}^{m} \beta_1^{k-j} \sum_{i=1}^d \frac{\nabla_i F(\widetilde w_t)^2 }{\sqrt{9G^2+\epsilon}}\nonumber\\
&=& \frac{\|\nabla F(\widetilde w_t)\|_2^2}{\sqrt{9G^2 + \epsilon}} \sum_{j=1}^{m}\sum_{k=j}^{m}\beta_1^{k-j}\nonumber\\
&\geq& \frac{\|\nabla F(\widetilde w_t)\|_2^2}{\sqrt{9G^2 + \epsilon}} \sum_{j=1}^{m}1\nonumber\\
&=& \frac{1}{\sqrt{9G^2 + \epsilon}} m \left\|\nabla F(\widetilde w_t)\right\|_2^2.
\end{eqnarray}
\textbf{\emph{Bounding $T_3$:}} We obtain
\begin{eqnarray}
\label{T3}
T_3 &\leq& \left(\sum_{k=1}^{m} \left\|\left(V_k^{(t)}\right)^{-1/2} \widetilde m_k^{(t)}\right\|_2\right)^2 \nonumber\\
&=&\left( \sum_{k=1}^{m}\frac{1}{1-\beta_1^k}\sqrt{\sum_{i=1}^d \frac{\left(m_{k,i}^{(t)}\right)^{2}}{v_{k,i}^{(t)}+\epsilon} }\right)^2\nonumber\\
&\leq& \left( \sum_{k=1}^{m} \frac{1}{\sqrt{\epsilon}} \left\|m_k^{(t)}\right\|_2\right)^2 \leq \left( \sum_{k=1}^{m}\frac{3G}{\sqrt{\epsilon}}\right)^2 \leq \frac{9G^2m^2}{\epsilon}.
\end{eqnarray}
In summary, we get
\begin{eqnarray}
\label{BoundDifference}
F\left( w^{(t)}_{m+1}\right) &\leq &F(\widetilde w_t)- \alpha_t \frac{m(1-\beta_1)}{\sqrt{9G^2+\epsilon}} \left\|\nabla F(\widetilde w_t)\right\|_2^2 + \frac{3G^2Lm(m-1)/(1-\beta_1)^2+9G^2Lm^2}{2\epsilon} \alpha_t^2 \nonumber\\
&=& F(\widetilde w_t) - Q_8\alpha_t m \left\|\nabla F(\widetilde w_t)\right\|_2^2 + Q_9 \alpha_t^2,\nonumber
\end{eqnarray}
where
\begin{eqnarray}
Q_8 &=& \frac{1-\beta_1}{\sqrt{9G^2+\epsilon}}\nonumber\\
Q_9 &=& \frac{3G^2Lm(m-1)/(1-\beta_1)^2+9G^2Lm^2}{2\epsilon}.\nonumber
\end{eqnarray}
\textbf{Proof of Theorem~\ref{thm:VRStrCvx}:} 

From Lemma~\ref{lem:BoundGradient}, we obtain the upper bound for each sample loss at each step. For any $1\leq n\leq N$, $1\leq k\leq m$ and $1\leq t\leq T$, we have
\begin{eqnarray}
    \left\|\nabla f_n\left(w_k^{(t)}\right)\right\|_2 \leq G.\nonumber
\end{eqnarray}
By applying this result to Lemma~\ref{lem:BoundStepVR}, we obtain
\begin{eqnarray}
    F\left(w^{(t)}_{m+1}\right) \leq F\left( \widetilde w_t\right) - Q_8\alpha_t m\left\|\nabla F(\widetilde w_t)\right\|_2^2 + Q_9 \alpha_t^2.\nonumber
\end{eqnarray}
According to Lemma~\ref{lem:ProjectionDecrease}, we have 
\begin{eqnarray}
    F\left(\widetilde w_{t+1}\right) \leq F\left( \widetilde w_t\right) - Q_8\alpha_t m\left\|\nabla F(\widetilde w_t)\right\|_2^2 + Q_9 \alpha_t^2.\nonumber
\end{eqnarray}
\color{black}
$F(w)$ is $c$-strongly convex, so we have
\begin{eqnarray}
\left\|\nabla F(w)\right\|_2^2 \geq 2c\left(F(w) - F^*\right),\nonumber
\end{eqnarray}
and thus,
\begin{eqnarray}
F(\widetilde w_{t+1}) \leq F(\widetilde w_t) - 2cQ_8\alpha_t m \left(F(w_t)-F^*\right) + Q_9 \alpha_t^2\nonumber,
\end{eqnarray}
which is equivalent to
\begin{eqnarray}
F(\widetilde w_{t+1}) - F^* \leq \left( 1 - \frac{C_2 m \alpha}{t}\right)\left(F(\widetilde w_t) - F^*\right) + Q_9 \alpha_t^2 \nonumber.
\end{eqnarray}
We obtain recursively
\begin{eqnarray}
F(\widetilde w_T) - F^* \leq \prod_{t=1}^{T-1}\left(1-\frac{C_2m\alpha}{t}\right) (F(\widetilde w_1) - F^*) + \sum_{t=1}^{T-1}\alpha_t \prod_{j=t+1}^{T-1} \left(1-\frac{C_2m\alpha}{j}\right). \nonumber
\end{eqnarray}
By definition, we have $C_2m\alpha<1$, and thus we can use Lemma~\ref{lem:Recursive} to obtain
\begin{equation}
    F(\widetilde w_T) - F^* \leq \OO\left(T^{-C_2m\alpha}\right).\nonumber
\end{equation}

\textbf{Proof of Theorem~\ref{thm:VRNonCvx}:} 
Applying Lemma~\ref{lem:BoundStepVR}, we have
\begin{eqnarray}
F(\widetilde w_{t+1}) - F(\widetilde w_{t})=F\left(\widetilde w^{(t)}_{m+1}\right) - F(\widetilde w_{t}) \leq - \alpha_t Q_8 m\left\|\nabla F(\tilde w_t)\right\|_2^2 + Q_9 \alpha_t^2.\nonumber
\end{eqnarray}
Then it follows
\begin{eqnarray}
F_{\inf} - F(\widetilde w_{1})&\leq& F(\widetilde w_{T+1}) - F(\widetilde w_{1}) \nonumber\\
&\leq& \sum_{t=1}^{T} -Q_8 m \alpha_t \left\|\nabla F(\widetilde w_t)\right\|_2^2 + Q_9\sum_{t=1}^T \alpha_t^2,\nonumber
\end{eqnarray}
and thus
\begin{eqnarray}
  \EE\left[ \left\|\nabla F(\widetilde w_\tau)\right\|_2^2\right] = \frac{ \sum_{t=1}^T \alpha_t \left\|\nabla F(\widetilde w_t)\right\|_2^2 }{\sum_{t=1}^T \alpha_t} \leq \frac{1}{\sum_{t=1}^T \alpha_t} \left\{\frac{F(\widetilde w_{1}) - F_{\inf}}{Q_8m} + \frac{Q_9}{Q_8m}\sum_{t=1}^T\alpha_t^2\right\}\nonumber.
\end{eqnarray}
This completes the proof.

\subsection{Proof of Theorem~\ref{thm:OnlineVRStrCvx}}
Similar to the proof in the previous section, we start by showing the following lemma.

\begin{lemma}
\label{lem:BoundStepOVR}
Given (1) and (3) in Assumption~\ref{assmpt:LSmoothGrad}, and by assuming that $\left\|\nabla f_n\left(w_k^{(t)}\right)\right\|_2\leq G$ holds for all $n,k$ and $t$, while setting $\alpha_k^{(t)}=\alpha_{t}\gamma^{m-k}$ for some $\beta_1<\gamma<1$, there exist positive constants $Q_{10}, Q_{11}$ and $Q_{12}$ such that Online VRADAM satisfies that for any $t$, 
\begin{equation*}
\EE_t F\left( w^{(t)}_{m+1} \right) - F(\widetilde w_t) \leq - Q_{10} \alpha_t \left\|\nabla F(\widetilde w_t)\right\|_2^2 + Q_{11} \alpha_t^2 +\frac{Q_{12}}{m} \alpha_t,
\end{equation*}
where $\EE_t[\cdot]$ denotes for expectation conditioned on $\widetilde w_t$.
\end{lemma}

Same as in the proof to Lemma\ref{lem:BoundStepVR}, we start by employing $L$-smoothness of gradient of $F(w)$ as follows
\begin{eqnarray}F({w}_{m+1}^{(t)}) 
&\leq& F(\widetilde{w}_t) + \nabla F(\widetilde{w}_t) ^\top \left({w}_{m+1}^{(t)} - \widetilde{w}_t\right) + \frac{L}{2}\left\|{w}_{m+1}^{(t)} - \widetilde{w}_t\right\|_2^2 \nonumber\\
&=& F(\widetilde{w}_t) + \nabla F(\widetilde{w}_t) ^\top \sum_{k=1}^{m} \left(w^{(t)}_{k+1} - w^{(t)}_{k}\right) + \frac{L}{2}\left\|\sum_{k=1}^{m} \left(w^{(t)}_{k+1} - w^{(t)}_{k}\right)\right\|_2^2\nonumber \\
&=& F(\widetilde w_{t}) - \nabla F(\widetilde w_t)^\top\sum_{k=1}^{m} \alpha_k^{(t)}\left(V_k^{(t)}\right)^{-1/2} \widetilde m_k^{(t)} + \frac{\alpha_t^2L}{2}\left\|\sum_{k=1}^{m}\left(V_k^{(t)}\right)^{-1/2} \widetilde m_k^{(t)}\right\|_2^2.\nonumber
\end{eqnarray}
By definition and the resetting option, we have
\begin{eqnarray}
\widetilde m_k^{(t)} = \frac{1-\beta_1}{1-\beta_1^k} \left(V_k^{(t)}\right)^{-1/2} \sum_{j=1}^k \beta_1^{k-j} g_j^{(t)}\nonumber,
\end{eqnarray}
and thus
\begin{eqnarray}
F\left( w^{(t)}_{m+1}\right)&\leq& F(\widetilde w_t) - (1-\beta_1) \nabla F(\widetilde w_t)^{\top}\sum_{k=1}^{m}\alpha_k^{(t)}\frac{1}{1-\beta_1^{k}} \left(V_k^{(t)}\right)^{-1/2} \sum_{j=1}^k \beta_1^{k-j} g_j^{(t)}\nonumber\\
&&+ \frac{\alpha_t^2L}{2}\left\|\sum_{k=1}^{m}\left(V_k^{(t)}\right)^{-1/2} \widetilde m_k^{(t)}\right\|_2^2\nonumber\\ 
&=&  F(\widetilde w_t) -  (1-\beta_1)\nabla F(\widetilde w_t)^{\top}\sum_{j=1}^{m} \left(\sum_{k=j}^{m} \alpha_k^{(t)}\frac{\beta_1^{k-j}}{1-\beta_1^k} \left(V_k^{(t)}\right)^{-1/2}\right) g_j^{(t)} 
\nonumber\\
&& +  \frac{\alpha_t^2L}{2}\left\|\sum_{k=1}^{m} \left(V_k^{(t)}\right)^{-1/2} \widetilde m_k^{(t)}\right\|_2^2\nonumber\\
&=& F(\widetilde w_t) - \alpha_t(1-\beta_1)\nonumber\\
&&\times \underbrace{\nabla F(\widetilde w_t)^{\top}\sum_{j=1}^{m} \left(\sum_{k=j}^{m} \gamma^{m-k}\frac{\beta_1^{k-j}}{1-\beta_1^k} \left(V_k^{(t)}\right)^{-1/2}\right) \left(\nabla F^{\BB^{(t)}_j}\left(w_j^{(t)}\right) - \nabla F^{\BB^{(t)}_j} \left(\widetilde w_t\right)\right)}_{T_1} \nonumber\\
&& - \alpha_t(1-\beta_1)\underbrace{\nabla F(\widetilde w_t)^{\top}\sum_{j=1}^{m}\left( \sum_{k=j}^{m}\gamma^{m-k} \frac{\beta_1^{k-j}}{1-\beta_1^k} \left(V_k^{(t)}\right)^{-1/2}\right)\left( \frac{1}{j}\sum_{l=1}^{j} \nabla F^{\BB_{l}^{(t)}}(\widetilde w_t)\right) }_{T_2}\nonumber\\
&& +  \frac{\alpha_t^2L}{2}\underbrace{\left\|\sum_{k=1}^{m} \left(V_k^{(t)}\right)^{-1/2} \widetilde m_k^{(t)}\right\|_2^2}_{T_3}.\nonumber
\end{eqnarray}
Term $T_3$ stays unchanged from (\ref{T3}) in the proof of Lemma~\ref{lem:BoundStepVR} and thus
\begin{eqnarray}
T_3\leq \frac{9G^2m^2}{\epsilon}. \nonumber
\end{eqnarray}

Term $T_1$ remains unchanged except for the following.
\begin{eqnarray}
    T_1 &\geq& -  \sum_{j=1}^{m} \left\| \sum_{k=j}^{m} \gamma^{m-k}\frac{\beta_1^{k-j}}{1-\beta_1^k} \left(V_k^{(t)}\right)^{-1/2}\nabla F(\widetilde w_t)\right\|_2\left\|\nabla F^{\BB^{(t)}_j}\left(w_j^{(t)}\right) - F^{\BB^{(t)}_j}(\widetilde w_t)\right\|_2\nonumber\\
    &\geq& -  \sum_{j=1}^{m} \left( \sum_{k=j}^{m} \gamma^{m-k}\frac{\beta_1^{k-j}}{1-\beta_1^k} \left\|\left(V_k^{(t)}\right)^{-1/2}\nabla F(\widetilde w_t)\right\|_2\right)\left\|\nabla F^{\BB^{(t)}_j}\left(w_j^{(t)}\right) - F^{\BB^{(t)}_j}(\widetilde w_t)\right\|_2\nonumber\\
    &\geq& -  \sum_{j=1}^{m} \left( \sum_{k=j}^{m} \frac{\beta_1^{k-j}}{1-\beta_1^k} \left\|\left(V_k^{(t)}\right)^{-1/2}\nabla F(\widetilde w_t)\right\|_2\right)\left\|\nabla F^{\BB^{(t)}_j}\left(w_j^{(t)}\right) - F^{\BB^{(t)}_j}(\widetilde w_t)\right\|_2,\nonumber\\
    &\geq& - \frac{3G^2L}{2(1-\beta_1)^3\epsilon} m(m-1) \alpha_t.\nonumber
\end{eqnarray}

Now we bound $T_2$.
Noticing that for all $a,b\in\RR^d$, we have $a^\top b=\frac{-\|a-b\|_2^2+\|a\|_2^2 + \|b\|_2^2}{2}\geq \frac{-\|a-b\|_2^2+\|a\|_2^2}{2}$, we let
\begin{eqnarray}
    G_{j}^{(t)}=\frac{1}{j}\sum_{l=1}^j\nabla F^{\BB_{l}^{(t)}}(\widetilde w_t),\nonumber
\end{eqnarray}
and then $T_2$ reads
\begin{eqnarray}
    T_2 &=&  \sum_{j=1}^m \sum_{k=j}^m \gamma^{m-k}\frac{\beta_1^{k-j}}{1-\beta_1^k} \nabla F(\widetilde w_t)^\top \left(V_{k}^{(t)}\right)^{-1/2}  G_j^{(t)}\nonumber\\
    &=& \sum_{j=1}^m \sum_{k=j}^m\gamma^{m-k} \frac{\beta_1^{k-j}}{1-\beta_1^k}\left(\left(V_{k}^{(t)}\right)^{-1/4} \nabla F(\widetilde w_t)\right)^\top \left(\left(V_{k}^{(t)}\right)^{-1/4}  G_j^{(t)}\right)\nonumber\\
    &\geq& - \underbrace{\sum_{j=1}^m \sum_{k=j}^m \gamma^{m-k}\frac{\beta_1^{k-j}}{2(1-\beta_1^k)}\left\|\left(V_{k}^{(t)}\right)^{-1/4} \left(\nabla F(\widetilde w_t)-G_j^{(t)}\right)\right\|_2^2}_{T_{21}}\nonumber\\
    &&+ \underbrace{\sum_{j=1}^m \sum_{k=j}^m \gamma^{m-k}\frac{\beta_1^{k-j}}{2(1-\beta_1^k)}\left\|\left(V_{k}^{(t)}\right)^{-1/4} \left(\nabla F(\widetilde w_t)\right)\right\|_2^2}_{T_{22}}.\nonumber
\end{eqnarray}
Term $T_{22}$ can be bounded in a similar manner as the proof to Lemma~\ref{lem:BoundStepVR}:
\begin{eqnarray}
% \label{T2}
T_{22} &=& \sum_{j=1}^{m}\sum_{k=j}^{m}\frac{\gamma^{m-k}}{2} \frac{\beta_1^{k-j}}{1-\beta_1^k} \nabla F(\widetilde w_t)^{\top} \left(V_k^{(t)}\right)^{-1/2} \nabla F(\widetilde w_t)\nonumber\\
&=&  \sum_{j=1}^{m}\sum_{k=j}^{m}\frac{\gamma^{m-k}}{2} \frac{\beta_1^{k-j}}{1-\beta_1^k} \sum_{i=1}^d \frac{\nabla_i F(\widetilde w_t)^2}{\sqrt{v_{k,i}^{(t)} + \epsilon}}\nonumber\\
&\geq& \sum_{j=1}^{m}\sum_{k=j}^{m} \frac{\gamma^{m-k}}{2}\beta_1^{k-j} \sum_{i=1}^d \frac{\nabla_i F(\widetilde w_t)^2 }{\sqrt{9G^2+\epsilon}}\nonumber\\
&=& \frac{\|\nabla F(\widetilde w_t)\|_2^2}{2\sqrt{9G^2 + \epsilon}} \sum_{j=1}^{m}\sum_{k=j}^{m}\gamma^{m-k}\beta_1^{k-j}\nonumber\\
&>& \frac{\|\nabla F(\widetilde w_t)\|_2^2}{2\sqrt{9G^2 + \epsilon}} \sum_{j=1}^m \sum_{k=j}^m\beta_1^{m-j}\nonumber\\
&=&\frac{\|\nabla F(\widetilde w_t)\|_2^2}{2\sqrt{9G^2 + \epsilon}} \sum_{j=1}^m (m-j)\beta_1^{m-j}>\frac{\beta_1\|\nabla F(\widetilde w_t)\|_2^2}{2\sqrt{9G^2 + \epsilon}},\nonumber
\end{eqnarray}
where the second inequality holds because $\gamma > \beta_1$ and the third inequality holds due to
$$\sum_{j=1}^m (m-j)\beta_1^{m-j} = \sum_{j=1}^{m-1} j\beta_1^j = \beta_1 + 2\beta_1^2+\cdots>\beta_1.$$

Now we bound $T_{21}$. According to the assumed boundary of sample gradients, the dispersion of gradients are uniformly bounded,
\begin{eqnarray}
    \EE_t\left[\left\|\nabla F^{\BB_{k}^{(t)}}(\widetilde w_t) -\nabla F(\widetilde w_t)\right\|_2^2\right]\leq 4G^2,\nonumber
\end{eqnarray}
for all $k$. In fact we have
\begin{eqnarray}
    \EE_t \left\|\nabla F(\widetilde w_t) - G_j^{(t)}\right\|_2^2\nonumber &=& \frac{1}{j^2}\EE_t  \left\|\sum_{l=1}^j \left(\nabla F(\widetilde w_t)-\nabla F^{\BB_{l}^{(t)}}(\widetilde w_t)\right)\right\|_2^2\nonumber\\
    &=& \frac{1}{j^2} \sum_l^j \EE_t \left\| \nabla F(\widetilde w_t) - \nabla F^{\BB_l^{(t)}}(\widetilde w_t)\right\|_2^2\leq \frac{4G^2}{j}.\nonumber
\end{eqnarray}
Then, the expectation of  $T_{21}$ is bounded by
\begin{eqnarray}
    \EE_t T_{21} &\leq& \frac{4G^2}{2\sqrt{\epsilon}(1-\beta_1)}\sum_{j=1}^m \sum_{k=j}^m \frac{\gamma ^{m-k}\beta_1^{k-j}}{j}\nonumber\\
    &<& \frac{4G^2}{2\sqrt{\epsilon}(1-\beta_1)}\sum_{j=1}^m \sum_{k=j}^m \frac{\gamma^{m-j}}{j} \nonumber\\
    &=&\frac{2G^2}{\sqrt{\epsilon}(1-\beta_1)}\sum_{j=1}^m \frac{\gamma^{m-j}(m-j)}{j}.\nonumber
\end{eqnarray}
We notice
\begin{eqnarray}
    \sum_{j=1}^m \frac{\gamma^{m-j}(m-j)}{j} &=& \sum_{j=1}^{m-1} \frac{ \gamma ^j j}{m-j} \nonumber\\
    &=&\sum_{j=1}^{\lfloor{(m-1)/2}\rfloor}\frac{ \gamma ^j j}{m-j} + \sum_{j=\lfloor{(m-1)/2}\rfloor}^{m-1}\frac{ \gamma ^j j}{m-j}\nonumber\\
    &\leq& \frac{1}{m-m/2} \sum_{j=1}^{\lfloor{(m-1)/2}\rfloor} \gamma^j  j + \frac{m-1}{m-(m-1)}\sum_{j=\lfloor{(m-1)/2}\rfloor}^{m-1}\gamma ^j\nonumber\\
    &\leq& \frac{2}{m} \sum_{j=1}^\infty \gamma^j j + (m-1) \sum_{j=\lfloor{(m-1)/2}\rfloor}^{\infty}\gamma ^j\nonumber\\
    &=& \frac{2}{m} \frac{\gamma}{(1-\gamma)^2} + (m-1) \frac{\gamma^{\lfloor{(m-1)/2}\rfloor}}{1-\gamma}.\nonumber
\end{eqnarray}
If $f(x)=x^2\gamma^{x/2}$ then $f'(x) = x^2\gamma^{x/2}(2/x+\log\gamma/2)$. Noting that $f'(-4/\log\gamma)=0$, the global maximum of $f$ is
$f(-4/\log\gamma) = 16/(e^2\log^2\gamma)$. Therefore we have
\begin{eqnarray}
    (m-1)\gamma^{\lfloor{(m+1)/2}\rfloor} \leq \frac{1}{m}(m^2\gamma^{m/2})\leq \frac{16}{e^2\log^2\gamma} \frac{1}{m},\nonumber
\end{eqnarray}
and then we have
\begin{eqnarray}
    \sum_{j=1}^m\frac{\gamma^{m-j}(m-j)}{j} \leq \left( \frac{2\gamma}{(1-\gamma)^2} + \frac{16}{e^2(1-\gamma)\log^2\gamma}\right)\frac{1}{m}.\nonumber
\end{eqnarray}
The upper bound of $\EE_t T_{21}$ reads
\begin{eqnarray}
    \EE_t T_{21}\leq\frac{2G^2}{\sqrt{\epsilon}(1-\beta_1)}\left( \frac{2\gamma}{(1-\gamma)^2} + \frac{16}{e^2(1-\gamma)\log^2\gamma}\right)\frac{1}{m},\nonumber
\end{eqnarray}
To summarize, we obtain
\begin{equation*}
\EE_t F\left( w^{(t)}_{m+1} \right) -  F(\widetilde w_t) \leq - Q_{10} \alpha_t \left\|\nabla F(\widetilde w_t)\right\|_2^2 + Q_{11} \alpha_t^2 +\frac{Q_{12}}{m}\alpha_t,
\end{equation*}
where
\begin{eqnarray}
    Q_{10} &=&    \frac{\beta_1(1-\beta_1)}{2\sqrt{9G^2+\epsilon}} \nonumber\\
    Q_{11} &=& \frac{3G^2Lm(m-1)}{2(1-\beta_1)^2\epsilon}+\frac{9G^2Lm^2}{2\epsilon}\nonumber\\
    Q_{12} &=& \frac{2G^2}{\sqrt{\epsilon}}\left( \frac{2\gamma}{(1-\gamma)^2} + \frac{16}{e^2(1-\gamma)\log^2\gamma}\right).\nonumber
\end{eqnarray}

\textbf{Proof of Theorem~\ref{thm:OnlineVRStrCvx}:} 

From Lemma~\ref{lem:BoundGradient}, we obtain the upper bound for each sample loss at each step. For any $1\leq n\leq N$, $1\leq k\leq m$ and $1\leq t\leq T$, we have
\begin{eqnarray}
    \left\|\nabla f_n\left(w_k^{(t)}\right)\right\|_2 \leq G.\nonumber
\end{eqnarray}
By applying this result to Lemma~\ref{lem:BoundStepOVR}, we obtain
\begin{eqnarray}
    \EE_t F\left(w^{(t)}_{m+1}\right) \leq  F\left( \widetilde w_t\right) - Q_{10}\alpha_t \left\|\nabla F(\widetilde w_t)\right\|_2^2 + Q_{11} \alpha_t^2 + \frac{Q_{12}}{m}\alpha_t.\nonumber
\end{eqnarray}
According to Lemma~\ref{lem:ProjectionDecrease}, we have that $F(\widetilde w_{t+1}) \leq F(w_{m+1}^{(t)})$ almost surely, therefore 
\begin{eqnarray}
   \EE_t F\left(\widetilde w_{t+1}\right) \leq F\left( \widetilde w_t\right) - Q_{10}\alpha_t \left\|\nabla F(\widetilde w_t)\right\|_2^2 + Q_{11} \alpha_t^2 + \frac{Q_{12}}{m}\alpha_t.\nonumber
\end{eqnarray}

Function $F(w)$ is $c$-strongly convex, so we have
\begin{eqnarray}
\left\|\nabla F(w)\right\|_2^2 \geq 2c\left(F(w) - F^*\right),\nonumber
\end{eqnarray}
and thus,
\begin{eqnarray}
\EE_t F(\widetilde w_{t+1}) \leq F(\widetilde w_t) - 2cQ_{10}\alpha_t \left(F(w_t)-F^*\right) + Q_{11} \alpha_t^2 + \frac{Q_{12}}{m}\alpha_t\nonumber,
\end{eqnarray}
which is equivalent to
\begin{eqnarray}
\EE_t F(\widetilde w_{t+1}) - F^* \leq \left( 1 - \frac{C_3 \alpha}{t}\right)\left(F(\widetilde w_t) - F^*\right) + Q_{11} \alpha_t^2  + \frac{Q_{12}}{m}\alpha_t\nonumber.
\end{eqnarray}
Taking full expectation on both sides we obtain recursively
\begin{eqnarray}
\EE F(\widetilde w_T) - F^* \leq \prod_{t=1}^{T-1}\left(1-\frac{C_3\alpha}{t}\right) (F(\widetilde w_1) - F^*) + \sum_{t=1}^{T-1}\frac{Q_{11}\alpha^2}{t^2} \prod_{j=t+1}^{T-1} \left(1-\frac{C_3\alpha}{j}\right) + \frac{Q_{12} }{m}\sum_{t=1}^{T-1}\frac{\alpha}{t} \prod_{j=t+1}^{T-1} \left(1-\frac{C_3\alpha}{j}\right) . \nonumber
\end{eqnarray}
By definition, we have $C_3\alpha<1$. We use Lemma~\ref{lem:Recursive} to obtain
that the first two terms are $\OO(T^{-C_3\alpha})$. The third term reads
\begin{eqnarray}
\sum_{t=1}^{T-1}\frac{1}{t}\prod_{j=t+1}^{T-1} \left(1-\frac{C_3\alpha}{j}\right)\leq \sum_{t=1}^{T-1}\frac{1}{t}\left(\frac{T}{t+1}\right)^{-C_3\alpha} < +\infty,\nonumber
\end{eqnarray}
where the first inequality uses (\ref{eq:boundprod}). We then obtain
\begin{eqnarray}
    \EE F(\widetilde w_T) - F^* \leq \OO(T^{-C_3\alpha }+ 1/m),\nonumber
\end{eqnarray}
which finishes the proof.
\color{black}

\section{Upper Bounds of the Optimal Solutions for Common Losses}
In order to properly choose the value of $M$ that satisfies Assumption~\ref{assmpt:LSmoothGrad}, the norm of optimal solution $\|w^*\|$ is needed. We show that for several common loss functions, $\|w^*\|_2$ can be estimated in linear complexity.
\subsection{Cross Entropy Loss with L2 Regularization}
\label{subsec:log}
Consider the logistic regression problem with feature matrix $X$ and labels $y$ where $X$ is an $N\times d$ matrix and $y\in\{0,1\}^N$. Let $X_i$ denote for the $i$th row of $X$. The full batch loss of the problem is
\begin{eqnarray}
    F_{\text{log}}(w) = \frac{1}{N} \sum_{i=1}^N\left[ y_i \log \left( \frac{1}{1+e^{-X_i^\top w}}\right) + (1-y_i) \log \left( 1-\frac{1}{1+e^{-X_i^\top w}}\right) \right]+ \frac{c}{2} \left\|w\right\|_2^2,
\end{eqnarray}
where $c>0$ is the regularization parameter.

If $w_{\text{log}}^* := \arg\min_{w} F_{\text{log}}(w)$, then
\begin{eqnarray}
    \left\|w^*_{\text{log}} \right\|_2&\leq& \frac{1}{c}\max_{1\leq i\leq N} \left\|X_i\right\|_2.\nonumber
\end{eqnarray}
The proof is as follows. By definition $w^*_{\text{log}}$ satisfies $\nabla F_{\text{log}}(w^*_{\text{log}}) = 0$, i.e.,
\begin{eqnarray}
    0 &=& \frac{1}{N} \sum_{i=1}^N\left[ y_i\frac{ e^{-X_i^\top w^*_{\text{log}}}}{1+e^{-X_i^\top w^*_{\text{log}}}} X_i - (1-y_i) \frac{e^{X_i^\top w^*_{\text{log}}}}{1+e^{X_i^\top w^*_{\text{log}}}}X_i\right] + cw^*_{\text{log}}.\nonumber
\end{eqnarray}
Thus we obtain
\begin{eqnarray}
    \left\|w^*_{\text{log}} \right\|_2&=&\left\| \frac{1}{cN} \sum_{i=1}^N  \left[y_i\frac{ e^{-X_i^\top w^*_{\text{log}}}}{1+e^{-X_i^\top w^*_{\text{log}}}} X_i - (1-y_i) \frac{e^{X_i^\top w^*_{\text{log}}}}{1+e^{X_i^\top w^*_{\text{log}}}}X_i\right]\right\|_2 \nonumber\\
    &=& \frac{1}{cN} \sum_{i=1}^N \left|y_i\frac{1}{1+e^{X_i^\top w^*_{\text{log}}}} - (1-y_i) \frac{e^{X_i^\top w^*_{\text{log}}}}{1+e^{X_i^\top w^*_{\text{log}}}}\right| \left\|X_i\right\|_2 \nonumber\\
    &\leq& \frac{1}{cN} \sum_{i=1}^N \left\|X_i\right\|_2\nonumber\\
    &\leq& \frac{1}{c}\max_{1\leq i\leq N} \left\|X_i\right\|_2,\nonumber
\end{eqnarray}
where the first inequality holds because $y_i$ can only take the value of $0$ or $1$, thus $\left|y_i\frac{1}{1+e^{X_i^\top w^*_{\text{log}}}} - (1-y_i) \frac{e^{X_i^\top w^*_{\text{log}}}}{1+e^{X_i^\top w^*_{\text{log}}}}\right|\leq \max\left\{ \frac{1}{1+e^{X_i^\top w^*_{\text{log}}}}, \frac{e^{X_i^\top w^*_{\text{log}}}}{1+e^{X_i^\top w^*_{\text{log}}}}\right\}\leq 1$.

\subsection{Softened Hinge Loss with L2 Regularization}
The hinge loss with feature matrix $X\in \RR^{N\times d}$ and labels $y\in \{-1,1\}^N$ is
\begin{eqnarray}
    \frac{1}{N}\sum_{i=1}^N \max\left\{0, 1 - y_i X_i^\top w\right\}.\nonumber
\end{eqnarray}
This function is not differentiable, thus we consider a softened version of hinge loss
\begin{eqnarray}
    F_{\text{hinge}}(w) = \frac{1}{N} \sum_{i=1}^N \log \left(1+ e^{1-y_i X_i^\top w}\right) + \frac{c}{2}\left\|w\right\|_2^2,\nonumber
\end{eqnarray}
where $c>0$ is the regularization parameter. 

If $w^*_{\text{hinge}}:= \arg\min_w F_{\text{hinge}}(w)$, then
\begin{eqnarray}
    \left\|w^*_{\text{hinge}} \right\|_2&\leq& \frac{1}{c}\max_{1\leq i\leq N} \left\|X_i\right\|_2.\nonumber
\end{eqnarray}
By definition $\nabla F_{\text{hinge}} \left(w^*_{\text{hinge}}\right) =0$ and thus
\begin{eqnarray}
    0 = \frac{1}{N} \sum_{i=1}^N \frac{-y_i e^{1-y_iX_i^\top w^*_{\text{hinge}}} X_i}{1+e^{1-y_i X_i^\top w^*_{\text{hinge}}}} + cw^*_{\text{hinge}}.\nonumber
\end{eqnarray}
We obtain
\begin{eqnarray}
    \left\|w^*_{\text{hinge}}\right\|_2 &=& \frac{1}{cN} \left\|\sum_{i=1}^N \frac{-y_i e^{1-y_iX_i^\top w^*_{\text{hinge}}} X_i}{1+e^{1-y_i X_i^\top w^*_{\text{hinge}}}}\right\|_2\nonumber\\
    &\leq & \frac{1}{cN} \sum_{i=1}^N |y_i| \left|\frac{e^{1-y_iX_i^\top w^*_{\text{hinge}}}}{1+e^{1-y_i X_i^\top w^*_{\text{hinge}}}}\right|\left\|X_i\right\|_2\nonumber\\
    &\leq& \frac{1}{cN} \sum_{i=1}^N \left\|X_i\right\|_2\nonumber\\
    &\leq& \frac{1}{c}\max_{1\leq i \leq N} \left\|X_i\right\|_2,\nonumber
\end{eqnarray}
which completes the argument.

\subsection{Mean Square Error}
Consider the linear regression problem 
\begin{eqnarray}
    \min_{w\in\RR^d} F_{\text{linear}}(w):=\left\|Xw-y\right\|_2^2,\nonumber
\end{eqnarray}
where $X$ is an $N\times d$ feature matrix and $y$ is an $N$ dimensional label vector. When $\det \left(X^\top X\right)\neq 0$, the closed-form solution to this problem is known as
\begin{eqnarray}
    w_{\text{linear}}^*  = \left(X^\top X\right)^{-1} X^{\top} y.\nonumber
\end{eqnarray}
We estimate this optimal solution with bounds that can be calculated within $\OO(Nd)$ complexity.

Let $X$ be a rational matrix and $X_{ij} = p_{ij}/q_{ij}$, where $p_{ij},q_{ij}$ are integers for all $1\leq i\leq N$ and $1\leq j \leq d$. Without loss of generality, $q_{ij}>0$. Let $Q$ be the least common denominator of $q_{ij}$ for all $1\leq i\leq N$ and $1\leq j \leq d$. If $\det \left(X^\top X\right)\neq 0$, then $w_{\text{linear}}^*$ satisfies 
   \begin{eqnarray}
      \left\|w_{\text{linear}}^*\right\|_2 \leq 2^{d/2-1} Q^4 N^{3/2} \left(\max_{1\leq i\leq N} \left\|X_i\right\|_2\right)^{3/2} \left\|y\right\|_2.\nonumber
\end{eqnarray}

The argument is as follows. We have $X = \widetilde X/Q$ where $\widetilde X$ is an integer matrix. We can bound the optimal solution as
\begin{eqnarray}
     \left\|w_{\text{linear}}^* \right\|_2 &\leq& \left\|\left(X^\top X\right)^{-1}\right\|_2 \left\|X\right\|_2 \left\|y\right\|_2\nonumber\\
     &=& Q^2 \left\|\left(\widetilde X^\top \widetilde X\right)^{-1}\right\|_2 \left\|X\right\|_2 \left\|y\right\|_2\nonumber\\
     &=& \frac{Q^2}{\lambda_{\text{min}}\left(\widetilde X^\top \widetilde X\right)}\left\|X\right\|_2 \left\|y\right\|_2,\nonumber
\end{eqnarray}
where $\lambda_{\text{min}}\left(\widetilde X^\top \widetilde X\right)$ is the smallest eigenvalue of $\widetilde X^\top \widetilde X$. We employ the lower bound proposed in \cite{piazza2002upper}, stating that $\sigma_{\min} (A) \geq |\det A| /\left( 2^{d/2-1} \|A\|_F\right)$, where $A$ is a $d\times d$ matrix and $\sigma_{\min} (A)$ is its minimal singular value. Since $\widetilde X ^\top \widetilde X$ is positive definite and Hermitian, its eigenvalue equals to its singular value
\begin{eqnarray}
    \lambda_{\text{min}}\left(\widetilde X^\top \widetilde X\right) &=& \sigma_{\text{min}}\left(\widetilde X^\top \widetilde X\right) \nonumber\\
    &\geq& \frac{\left|\det \left(\widetilde X ^\top \widetilde X\right)\right|}{2^{d/2-1} \left\|\widetilde X^\top \widetilde X\right\|_F},\nonumber\\
    &\geq& \frac{1}{2^{d/2-1} \left\|\widetilde X^\top \widetilde X\right\|_F}\nonumber\\
    &=& \frac{1}{2^{d/2-1} Q^2 \left\| X^\top  X\right\|_F},\nonumber
\end{eqnarray}
where $\|\cdot\|_F$ is the Frobinius norm. The second inequality holds because $\left|\det\left(\widetilde X ^\top \widetilde X\right)\right|$ is a positive integer, and therefore $\left|\det\left(\widetilde X ^\top \widetilde X\right)\right|\geq 1$. Connecting the above yields
\begin{eqnarray}
    \left\|w_{\text{linear}}^*\right\|_2 &\leq& 2^{d/2 - 1} Q^4 \left\|X^\top X\right\|_F \left\|X\right\|_2 \left\|y\right\|_2\nonumber\\
    & \leq& 2^{d/2 - 1} Q^4 \left\|X^\top X\right\|_F \left\|X\right\|_F \left\|y\right\|_2\nonumber\\
    &\leq& 2^{d/2 - 1} Q^4 \left\|X\right\|_F^3 \left\|y\right\|_2.\nonumber
\end{eqnarray}
The last inequality $\left\|X^{\top} X\right\|_F \leq \left\| X\right\|_F^2$ is well known.
Noticing that
\begin{eqnarray}
    \left\|X\right\|_F = \sqrt{\sum_{i=1}^N\sum_{j=1}^d X_{ij}^2}
    = \sqrt{\sum_{i=1}^N \left\|X_i\right\|_2^2} 
    \leq \sqrt{N} \max_{1 \leq i\leq N} \left\|X_i\right\|_2,\nonumber
\end{eqnarray}
we obtain
\begin{eqnarray}
      \left\|w_{\text{linear}}^*\right\|_2 \leq 2^{d/2-1} Q^4 N^{3/2} \left(\max_{1\leq i\leq N} \left\|X_i\right\|_2\right)^{3/2} \left\|y\right\|_2.\nonumber
\end{eqnarray}

\section{Experiments}

\subsection{Network Structure}
\begin{table}[H]
    \centering
    \begin{tabular}{c|c|c|c}
    \hline
      Dataset   & Input dimension & Hidden dimension & Output Dimension\\\hline
        CovType & 98 & 100 & 7\\ 
        MNIST & 784 & 100 & 10\\ \hline
    \end{tabular}
    \caption{Feedforward network structure}
    \label{tab:FFNStructure}
\end{table}
The feedforward networks used in the experiments have two fully connected layers with the dimensions described in Table~\ref{tab:FFNStructure}.

The structure of the CNN used in the experiments is described as follows. The CNN is mainly composed of two convolution layers, two max pooling layers and one fully connected layer. The kernel size of the convolution layers is 4 and the kernel size of the pooling layers is 2. The numbers of channels of the two convolution layers are 16 and 32, respectively, and the dimensions of the fully connected layer are 32 for input and 10 for output.

\subsection{Additional Results}
\begin{figure}[H]
    \centering
    \begin{subfigure}[b]{0.3\textwidth}
        \includegraphics[width=\textwidth]{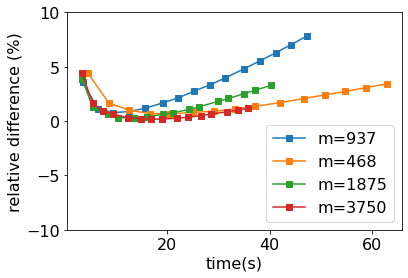}
        \caption{{ CIFAR10Emb with LogReg}}
    \end{subfigure}
    \hfill
    \begin{subfigure}[b]{0.3\textwidth}
        \includegraphics[width=\textwidth]{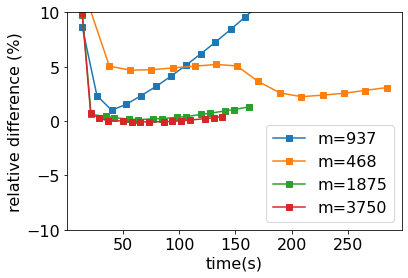}
        \caption{{ MNIST with CNN}}
    \end{subfigure}
    \hfill
    \begin{subfigure}[b]{0.3\textwidth}
        \includegraphics[width=\textwidth]{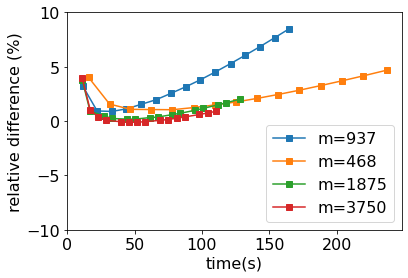}
        \caption{{MNIST with FFN}}
    \end{subfigure}
    \caption{Relative difference of training loss for VRADAM comparing with ADAM}
\end{figure}
\begin{figure}[H]
    \centering
    \begin{subfigure}[b]{0.3\textwidth}
        \includegraphics[width=\textwidth]{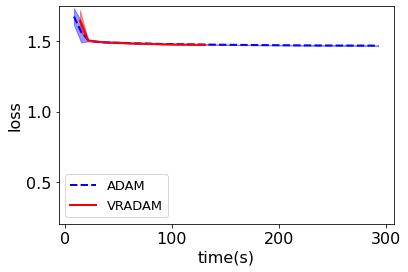}
        \caption{ MNIST with CNN}
    \end{subfigure}
    \hfill
    \begin{subfigure}[b]{0.3\textwidth}
        \includegraphics[width=\textwidth]{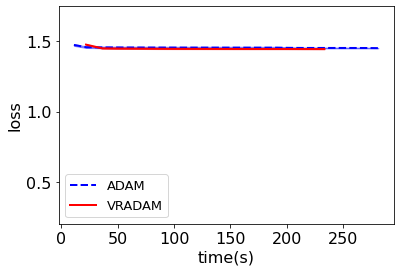}
        \caption{CovType with LogReg}
    \end{subfigure}
    \hfill
    \begin{subfigure}[b]{0.3\textwidth}
        \includegraphics[width=\textwidth]{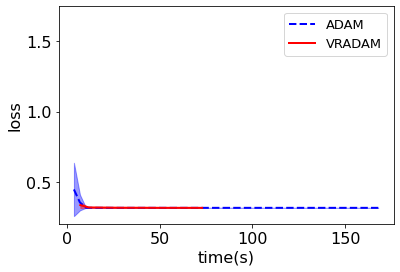}
        \caption{NSK-KDD with LogReg}
    \end{subfigure}
    \caption{Deviation of VRADAM and ADAM}
\end{figure}

\begin{figure}[H]
    \centering
    \begin{subfigure}[b]{0.3\textwidth}
        \includegraphics[width=\textwidth]{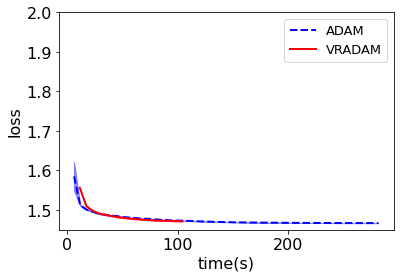}
        \caption{ MNIST with FFN}
    \end{subfigure}
    \hfill
    \begin{subfigure}[b]{0.3\textwidth}
        \includegraphics[width=\textwidth]{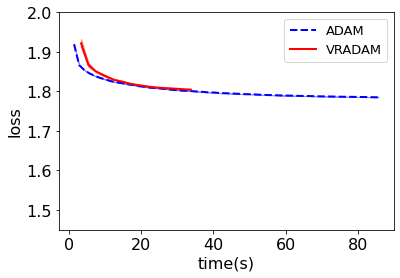}
        \caption{CIFAR10Emb with LogReg}
    \end{subfigure}
    \hfill
    \begin{subfigure}[b]{0.3\textwidth}
        \includegraphics[width=\textwidth]{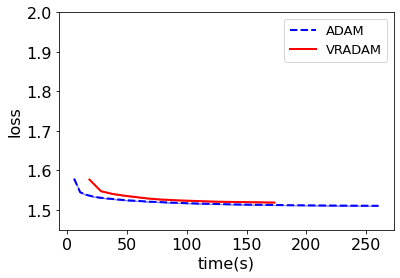}
        \caption{MNIST with LogReg}
    \end{subfigure}
    \caption{Deviation of VRADAM and ADAM. The values of deviation are close to zero therefore hard to be noticed from the plots.}
\end{figure}
\begin{figure}[H]
    \centering
    \begin{subfigure}[b]{0.3\textwidth}
        \includegraphics[width=\textwidth]{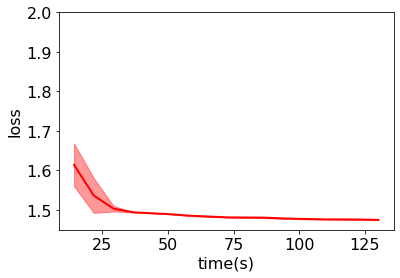}
        \caption{ MNIST with CNN}
    \end{subfigure}
    \hfill
    \begin{subfigure}[b]{0.3\textwidth}
        \includegraphics[width=\textwidth]{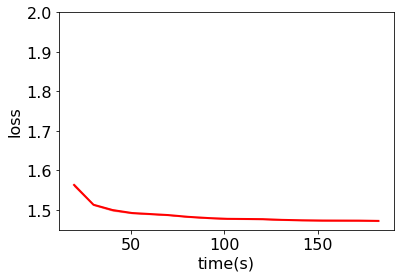}
        \caption{MNIST with FFN}
    \end{subfigure}
    \hfill
    \begin{subfigure}[b]{0.3\textwidth}
        \includegraphics[width=\textwidth]{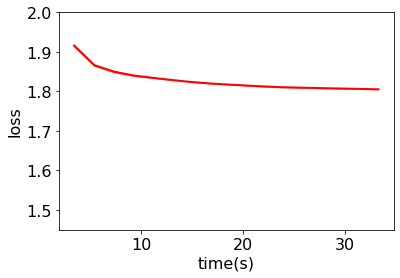}
        \caption{CIFAR10Emb with LogReg}
    \end{subfigure}
    \vfill
    \begin{subfigure}[b]{0.3\textwidth}
        \includegraphics[width=\textwidth]{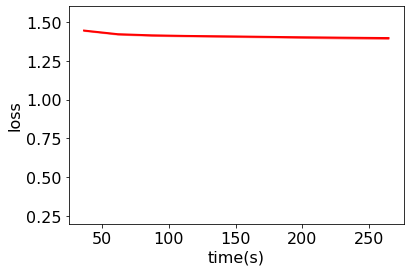}
        \caption{ CovType with LogReg}
    \end{subfigure}
    \hfill
    \begin{subfigure}[b]{0.3\textwidth}
        \includegraphics[width=\textwidth]{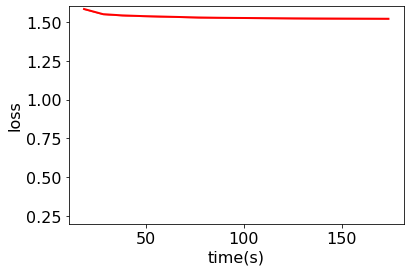}
        \caption{MNIST with LogReg}
    \end{subfigure}
    \hfill
    \begin{subfigure}[b]{0.3\textwidth}
        \includegraphics[width=\textwidth]{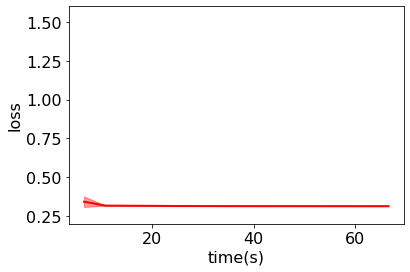}
        \caption{NSL-KDD with LogReg}
    \end{subfigure}
    \caption{Sensitivity on initial points for VRADAM}
\end{figure}
\color{black}
\end{document}